\documentclass[sigconf]{acmart}
\acmSubmissionID{1384}

\AtBeginDocument{%
  }

\copyrightyear{2026}
\acmYear{2026}
\setcopyright{cc}
\setcctype{by}
\acmConference[IDC '26]{Proceedings of the 25th Interaction Design and Children Conference}{June 22--25, 2026}{Brighton, United Kingdom}
\acmBooktitle{Proceedings of the 25th Interaction Design and Children Conference (IDC '26), June 22--25, 2026, Brighton, United Kingdom}
\acmDOI{10.1145/3773077.3806112}
\acmISBN{979-8-4007-2283-7/2026/06}

\usepackage{natbib}
\usepackage{xcolor}
\usepackage{soul}

\begin{document}

\title[Supporting FSP with Robot-Facilitated Home-Based Activities]{Supporting Family-School Partnerships with Robot-Facilitated Home-Based Activities}

\author{Michael F. Xu}
\orcid{0009-0000-2632-2428}
\affiliation{%
  \institution{University of Wisconsin--Madison}
  \city{Madison}
  \state{Wisconsin}
  \country{USA}
}
\email{michaelfxu@cs.wisc.edu}

\author{Qiyao Yang}
\orcid{0009-0007-5484-8938}
\affiliation{%
  \institution{University of Wisconsin--Madison}
  \city{Madison}
  \state{Wisconsin}
  \country{USA}
}
\email{qyang254@wisc.edu}

\author{Heather Kirkorian}
\orcid{0000-0002-7990-7777}
\affiliation{%
  \institution{University of Wisconsin--Madison}
  \city{Madison}
  \state{Wisconsin}
  \country{USA}
}
\email{kirkorian@wisc.edu}

\author{Bilge Mutlu}
\orcid{0000-0002-9456-1495}
\affiliation{%
  \institution{University of Wisconsin--Madison}
  \city{Madison}
  \state{Wisconsin}
  \country{USA}
}
\email{bilge@cs.wisc.edu}


\begin{abstract}
Family-school partnerships (FSP) are critical to children’s development, yet families often face barriers such as time constraints, fragmented communication, and limited opportunities for meaningful engagement. As a step toward facilitating broader family-school partnerships, we explore a novel approach that integrates a social robot into family settings, specifically supporting \textit{home-based} activities. Through interviews and co-design sessions, we designed and developed a robotic system informed by both parents and children, that supported, among other interactions, family communication about school topics. We evaluated the robot in a week-long, in-home study with 10 families. Our findings show how families integrated the robot into daily life, how parental facilitation styles shaped use, and how families perceived both the helpfulness and challenges of the robot. We contribute empirical insights, a modular system, and design implications for family- and child-robot interactions. We discuss ethical and privacy considerations, and broaden the design space for technologies supporting family-school partnerships.
\end{abstract}

\ccsdesc[300]{Human-centered computing~Empirical studies in HCI}
\ccsdesc[300]{Human-centered computing~Empirical studies in interaction design}
\ccsdesc[300]{Computer systems organization~Robotics}
\ccsdesc[100]{Human-centered computing~User studies}
\keywords{social robots, child-robot interaction, home-school communication, family-centered design, co-design, in-home evaluation}
\begin{teaserfigure}
  \includegraphics[width=\textwidth]{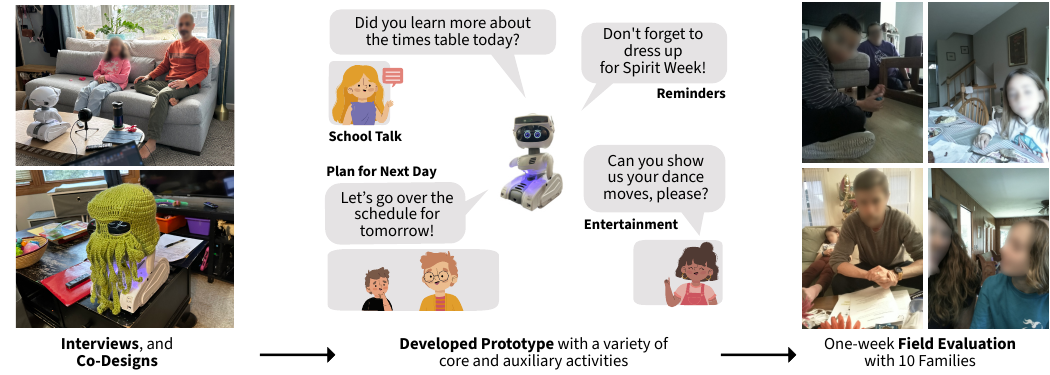}
  \caption{To systematically explore how social robots can facilitate home-based activities for school-related communication, we carried out three consecutive studies. Based on insights from Formative Study 1 (needs-finding interviews) and Formative Study 2 (co-design sessions), we iteratively designed and refined several \textit{Core Activities} (\textit{School Talk}, \textit{Plan for Next Day}, and \textit{Morning Rush}), along with several \textit{Auxiliary Features} such as reminders and entertainment features (\textit{e.g.}, jokes, dances, and casual conversations). We then deployed the prototype to ten families for a week-long in-home evaluation (Evaluation Study).}
  \Description[Study overview from co-design to family deployment]{Composite overview figure showing the study progression from early design work to an in-home field evaluation. On the left, two photos represent interviews and co-design activities with families in home settings. In the center, a mobile social robot is shown alongside four example interaction types developed for the prototype: School Talk, with a prompt asking whether the child learned more about multiplication; Plan for Next Day, with a prompt about reviewing tomorrow’s schedule; Reminders, with a prompt about dressing up for Spirit Week; and Entertainment, with a prompt asking the family to show dance moves. Small illustrated child characters accompany some of these example activities. Arrows connect the left, center, and right sections to indicate progression. On the right, four photos depict the one-week field evaluation conducted with 10 families in home environments.}
  \label{fig:teaser}
\end{teaserfigure}

\maketitle

\section{Introduction}

In the \textit{Handbook of School-Family Partnerships}, \citet{christenson2010handbook} started by recognizing the important role of families in a child's overall success:
\begin{quote}
    \textit{Although schools are the institution formally charged with educating our nation’s youth, the role of families and the home environment in students’ development and academic, behavioral, social, and emotional competence is undisputed.}
\end{quote}
They further argued that in addition to treating schools and families separately, what sits at the intersection of these two systems---the family-school partnership (FSP)---is also critical. Conceptual frameworks characterize FSP as multidimensional, encompassing both school-based involvement and home-based practices~\cite{epstein1995school, bronfenbrenner1979contexts, coleman1988social}. Empirical research further shows that effective partnerships are associated with improved academic performance, behavior, and social outcomes across diverse populations~\cite{henderson2002new}.

Despite these benefits, developing and sustaining effective FSP remains challenging. Families often face barriers such as limited time, knowledge, or access, while teachers and schools struggle to implement scalable and sustainable partnership practices~\cite{hornby2011barriers, epstein1995school}. Prior work suggests that home-based activities such as everyday conversations, routines, and parental support offer a promising and underexplored context for strengthening FSP without requiring major institutional change~\cite{henderson2007beyond, jeynes2005meta, jeynes2010salience,jeynes2007relationship}. In addition,  \citet{epstein1995school} argues that students generally ``want their families to be more knowledgeable partners about schooling and are willing to take active roles in assisting communications between home and school,'' but they need more support and guidance on ``how they can conduct important exchanges with their families about school activities, homework, and school decisions.''

To capitalize on this potential and address this gap, this paper focuses on \textbf{supporting FSP via home-based activities}, while setting the stage for broader FSP facilitation \textit{in future work}. In particular, this paper explores a novel approach that integrates a social robot into various home-based activities that support communications about school, while positioning the student as an active participant in the process. Building on this motivation, our work is guided by three research questions:  
\begin{itemize}
    \item \textbf{RQ1:} What types of home-based activities can a social robot facilitate to support school-related communications?
    \item \textbf{RQ2:} What are families’ preferences for the design and interaction patterns of such a robot?
    \item \textbf{RQ3:} How would families use and utilize such a social robot in their homes?  
\end{itemize}

To address these questions, we conducted a series of three studies: semi-structured interviews (Formative Study 1), co-design sessions (Formative Study 2), and a one-week in-home evaluative study (Evaluation Study) (Figure~\ref{fig:teaser}). We designed, developed, and deployed a system built around a social robot. Past work has demonstrated the potential of social robots to effectively engage with family members, especially the children, and facilitate interpersonal communication~\cite{chen2022designing, chen2025social, panicker2025haru, ho2023designing}. Leveraging those potentials, we conducted interviews and co-design sessions to better understand the needs, challenges, and preferences of both the parents and children, which we then translated into a variety of interactions and robot capabilities. From the pool of family-suggested activities, we selected and implemented both \textit{core activities} that directly support FSP, and \textit{auxiliary activities} that may have an indirect impact such as supporting engagement with the robot. An example of a \textit{core activity} is the ``School Talk'' activity, which prompted children to reflect on their school day, while parents may either directly participate in the conversation, or listen by being nearby. Examples of \textit{auxiliary activities} include dancing, telling jokes, and providing proactive reminders. We then deployed the system with 10 families over a week-long, in-home evaluation study. 

Synthesizing interview data with system usage logs, we report findings regarding the challenges families face, patterns of use,  design recommendations, and broader implications. We found that families perceived the robot-facilitated home-based activities as being helpful, both in facilitating school-related information exchange, and in reducing parents’ burden by offloading specific tasks such as getting ready for school and reminders about school events. 


The contributions of this work are four-fold:  
\begin{itemize}
    \item \textit{Design Space:} We examine how a social robot can support the \emph{home-based dimension} of family-school partnerships, articulating the design space and laying the groundwork for future work that more directly involves schools.
  \item \textit{Empirical:} Through a week-long, in-home field study, we investigate how families integrate and utilize a social robot within their day-to-day life and parent-child interactions.
  \item \textit{System:} We present a modular system framework that can support diverse family-robot activities and be adapted for other in-the-wild HRI studies.
  \item \textit{Practical:} We derive design implications for social robots that engage with families in home settings, accounting for family routines, parental facilitation styles, and multi-user household dynamics.
\end{itemize}

In the following sections, we review related work that informs our study, describe the series of studies conducted, and present our findings and their implications.

\section{Related Work}
In this section, we first review the role of home-based activities in the context of Family-School Partnerships, from both a theoretical and an empirical perspective. Next, we review relevant HCI work facilitating information flow between family and school. Finally, we describe the relevant work in the field of human-robot interaction, specifically those pertinent to child-robot interactions and family-robot interactions.

\subsection{Home-based Activities and FSP}
The importance of home-based activities in supporting family-school partnerships is well established across both theoretical and empirical literature. From a theoretical perspective, our work draws on Bronfenbrenner’s ecological model~\cite{bronfenbrenner1979contexts}, Epstein’s overlapping spheres of influence~\cite{epstein1987toward}, and social capital theory~\cite{bourdieu2018forms, coleman1988social, lareau1987social}, which together conceptualize FSP as a critical, multidimensional developmental context spanning both home and school. These perspectives are further operationalized by conceptual frameworks that articulate concrete parental involvement processes~\cite{epstein1995school, hoover1995parental, hoover1997parents, hoover2005final}, many of which are embedded in everyday home practices. Empirical research similarly underscores the importance of home-based activities. A large-scale meta-analysis of 77 family–school partnership interventions identified \textit{home-based involvement} as the only structural component consistently associated with positive effects across multiple child outcomes, including academic achievement, academic behavior, and social-behavioral competence~\cite{smith2020effects}. Together, these findings suggest that interventions situated within family contexts, particularly those that support communication and coordination with schools activities, may be especially effective. This motivates our exploration of how technology, such as an embodied social robot, can facilitate and scaffold home-based activities that contribute to family-school partnerships.

\subsection{Home-Based Technologies for Families}

Prior work has examined how technologies can support coordination and communication within the home. Researchers have studied shared family calendars, household scheduling practices, and reminder-oriented systems that help families manage routines, errands, and everyday responsibilities \cite{neustaedter2009calendar, plaisant2006shared, davidoff2010routine}. Related work in family informatics and CSCW has further shown that domestic coordination is not merely a logistical task, but is embedded in ongoing family relationships, communication practices, and collaborative care work \cite{pina2017personal, richards2022understanding}. Other systems have explored domestic communication and awareness more directly, for example through always-on video links that let families monitor one another's availability, exchange brief handwritten messages, and share everyday life across distance \cite{judge2010family}. More recently, HCI researchers have begun to examine conversational and voice-based technologies in domestic family settings. Studies of smart speakers and voice assistants show that these systems can become part of family communication and coordination, while also introducing challenges around breakdowns, role expectations, and differences in how adults and children engage with them \cite{beneteau2019communication, beneteau2020parenting, garg2020he, ammari2019music}.

Together, this literature suggests that technologies can play an important role in supporting home routines and family communication, but has focused less on embodied conversational systems designed specifically for \textit{school-related} family practices. In this work, we build on the strengths of technology for efficient and engaging communications, and explore the potential roles of an embodied robot. We foreground a family-centered design approach to account for perspectives from both the parents and the children, highlighting the role of the children in these interactions, while keeping the effort required of the parents in check.

\subsection{Human-Robot Interactions}
Since prior work examining robots in the context of family-school partnerships is sparse, we build on a broader body of research on child-robot interaction and the adoption of robots in family settings.

\subsubsection{Child-Robot Interactions}
Prior research has demonstrated the potential of embodied robots to engage children across a range of contexts, including eliciting information and disclosure through interviews and conversations~\cite{wood2013robot, bethel2016using, abbasi2022can}, sustaining engagement over time~\cite{ahmad2017adaptive, neerincx2021social, cagiltay2022understanding}, and supporting learning, relationship building, and social connection~\cite{belpaeme2018social, kanda2012children, gillet2020social}. For example, studies have shown that children may be more willing to self-disclose to robots than to human interviewers~\cite{bethel2016using}, and that robots can foster higher engagement compared to screen-based technologies in certain settings~\cite{neerincx2021social}. Prior reviews further suggest that children often perceive robots as social entities and may form relational bonds with them~\cite{van2020child}.

\subsubsection{Robots in Families}
Research on robots in family contexts highlights the complexity of multi-user environments. Although there is a general perception among parents that robots could be more helpful with child-centered family routines and tasks~\cite{xu2024robots}, it is also well-documented that family members often hold differing expectations and preferences for robotic systems~\cite{cagiltay2020investigating, de2016long, cagiltay2022understanding, he2025developing}. Such dynamics can pose challenges for adoption and sustained use, prompting calls for family-centered approaches to robot design~\cite{cagiltay2024toward}. Within this space, several studies have demonstrated the potential of social robots to facilitate intra-family communication. Prior work shows that robots can act as conversational catalysts between parents and children, enhancing the quality and frequency of interactions~\cite{chen2025social, ho2023designing}, supporting shared activities~\cite{kim2022can}, and reflecting family values and practices~\cite{panicker2025haru}.

Building on these prior works, our study explores a novel approach that integrates a social robot into home-based activities aimed at supporting school-related communications. We leverage the demonstrated potential of social robots to engage with family members -- particularly children -- and to act as facilitators of communication within the family unit. At the same time, we acknowledge the complexity of home environments and the influence of family dynamics on robot adoption and use. Accordingly, we adopt a family-centered approach to iteratively design and implement our research prototype, incorporating the perspectives of both parents and children throughout the process.

\section{Methods Overview}
To address our research questions, we iteratively conducted a series of three studies. RQ1 was explored through needs-finding interviews with families (Formative Study 1). RQ2 was examined through co-design sessions (Formative Study 2). RQ3 was investigated through a one-week in-home deployment of the social robot, during which we collected usage data and conducted post-study interviews (Evaluation Study).

\subsection{Study Design}
The following study design was approved by the author's Institutional Review Board. To systematically take on this underexplored space, we carried out three consecutive studies. In the first two formative studies, we aimed to understand the needs and preferences of the families, based on which we designed and implemented the interactions on the robot prototype. In the third and final study, we conducted a week-long field evaluation. Formative Study 1 was a semi-structured interview, focusing on the current state and challenges of the families related to school communication and engagement, and how the robot could potentially help. Formative Study 2, a co-design session, focused on families' design preferences for the activities and interaction capabilities implemented on the robot. Finally, in the Evaluation Study, we focused on assessing the real-world adoption of the robot, how it was used (or not used), and whether it showed \textit{potential} in facilitating home-based school communications. With a consideration for ecological validity, all three studies were conducted at the participants' homes.

\subsection{Study Timeline}
Across the three phases, data collection took place over approximately three months, from March 2025 to June 2025: Formative Study 1 was conducted in March and early April, Formative Study 2 in late April to early May, and the Evaluation Study in late May to mid-June. Although the participating children's schools followed different academic calendars, data collection occurred primarily during the late spring portion of the school year. We prepared four robot kits to facilitate parallel deployments; at any given time, no more than four families were actively participating in the Evaluation Study. Because the studies focused on school-related family routines, the timing of data collection may have influenced both the salience of particular school responsibilities and families’ engagement with the system. For example, for F13 and F14 the deployment concluded at the same time as the end of their school year, meaning their participation took place during the final week of school.

\subsection{Participants}
We recruited parents who had at least one child between age 9 and 12 (inclusive). Other children within the family who were outside of the target age range were encouraged to participate as well. Although families consented to and were paid for each study independently, all participants in later studies had taken part in earlier studies, and we did not recruit new families in later phases. Overall, 14 families enrolled (with a total of 27 children), and 10 families completed the entire three-study process. The four families who did not proceed either explicitly declined the invitation to continue onto the next study, or did not respond to further invitations. Among all four, only F2 explained to us that they are declining to proceed because they do not feel a need for additional support in this aspect. The children involved in these studies predominantly attended public schools in various local school districts. The majority of the households are considered middle- to upper-income (\textit{i.e.}, self-reported annual household incomes of \$100,000 USD or higher). Table \ref{tab:family_overview} summarizes the main participating family members across study phases, including children's ages/genders and the main family members who participated in each phase. Participation varied across families and study phases; some sessions involved only one parent, while others included both parents and one or more children. In the findings reported below, quotes from \textbf{parents} and \textbf{children} are prefixed with \textbf{P} and \textbf{C}, respectively.

    

\begin{table*}
  \centering
  \caption{Overview of family participation across study phases, including the children in each family and the main family members who participated in Formative Study 1 (FS1), Formative Study 2 (FS2), and the Evaluation Study (ES) interview. Child labels indicate age and gender (\textit{e.g.}, 9M, 11F). Because participation sometimes shifted during home visits, the listed participants represent the main contributors in each phase rather than an exhaustive record of everyone present.}
  \label{tab:family_overview}
  \begin{tabular*}{\textwidth}{@{\extracolsep{\fill}} l l c c c c @{}}
    \toprule
    \textbf{Family} \# & \textbf{Studies} & \textbf{Children} & \textbf{FS1} & \textbf{FS2} & \textbf{ES Exit Interview} \\
    \midrule
    F1  & FS1 & 9M, 5F  & Mom, 9M, 5F & -- & -- \\
    F2  & FS1 & 10M, 7F  & Dad, Mom, 10M & -- & -- \\
    F3  & FS1 & 12F  & Mom & -- & -- \\
    F4  & FS1 & 9F, 7F, 5F  & Mom & -- & -- \\
    F5  & FS1, FS2, ES & 10M, 9F  & Dad, 10M & Dad, 10M, 9F & Dad, Mom, 10M \\
    F6  & FS1, FS2, ES & 9F, 2M  & Dad, 9F & Dad, 9F & Dad, 9F \\
    F7  & FS1, FS2, ES & 9F, 5M  & Dad, 9F & Dad, 9F & Dad, 9F \\
    F8  & FS1, FS2, ES & 11F, 9F  & Dad, 11F, 9F & Dad, 11F, 9F & Dad, 11F, 9F \\
    F9  & FS1, FS2, ES & 11F, 8F  & Mom, 11F & Mom, 11F, 8F & Mom, 11F \\
    F10 & FS1, FS2, ES & 11F, 9M  & Mom, 9M & Mom, 9M & Mom, 9M \\
    F11 & FS1, FS2, ES & 12F, 10F, 8M  & Dad, Mom, 12F, 10F & Dad, Mom, 12F, 10F & Dad, Mom, 12F, 10F \\
    F12 & FS1, FS2, ES & 11F, 11F  & Mom, 11F, 11F & Mom, 11F, 11F & Dad, Mom, 11F, 11F \\
    F13 & FS1, FS2, ES & 10F  & Dad & Dad, 10F & Dad, 10F \\
    F14 & FS1, FS2, ES & 10F  & Dad, Mom, 10F & Dad, 10F & Dad, Mom, 10F \\
    \bottomrule
  \end{tabular*}
\end{table*}


\section{Formative Study 1}
Rather than directly connecting to school stakeholders in this prototype, we focused on the home-based routines and communication practices through which family-school partnership is enacted in everyday life. Formative Study 1 focused on exploring \textit{RQ1}, how families envision a social robot to facilitate their partnerships with schools. We first set out to understand the current state of family-school communications and the challenges the families face.

\subsection{Study Design and Procedure}

We conducted semi-structured interviews with the families. We started by learning about the current state of family-school partnerships, covering aspects such as the communication channels and the topics involved. We then transitioned into the challenges families face regarding these communications and collaborations with the schools and teachers, and mitigation strategies families have tried. Roughly halfway through each interview, we introduced the robot we had chosen for this study, the Misty II robot\footnote{\url{https://www.mistyrobotics.com/misty-ii}}, and brainstormed with the families about ways in which a robot like that may be able to support them. Example questions during the interview include: \textit{What are some of the challenges you are experiencing in terms of school communication and engagement?} and \textit{What are some features or capabilities that you think a robot like this could have, that will help you with the challenges you are facing?} Overall, the visit lasted approximately an hour for most families, ranging between 30 and 75 minutes. Families were paid $\$20$ USD per hour.

Interviews were conducted in families’ homes and at least one parent was required to participate. Although children were not required to participate, most families had some level of children participation (often running around, chiming in for a few minutes at times). Rather than interviewing family members separately, we conducted the session as a joint family conversation, directing questions to the group and following up with parents or children as appropriate. When children were present, they were addressed directly and invited to share their own experiences and ideas, especially during discussion of routines and robot design concepts. However, because the interview topics often centered on school communication, coordination, and mitigation strategies, parents were more often the primary respondents on these challenges. Consequently, the findings in this phase reflect family-level perspectives, but are weighted more heavily toward parents' accounts of communication-related difficulties.

\subsubsection{Data Collection and Analysis}
Interviews were audio recorded and later transcribed. Additional photos were used as contextual artifacts to document existing family strategies for managing school-related communication and coordination. The first author conducted a reflexive thematic analysis \cite{braun2006using, byrne2022worked, braun2019reflecting} focused on RQ1. Analysis began with repeated readings of the transcripts and accompanying notes and photos to become familiar with families' current practices, communication challenges, and ideas for robot support. The first author then generated inductive codes in a spreadsheet and iteratively refined them across interviews. Example early codes included \textit{communication channels} such as email, web portals, take-home flyers, and calendars. \textit{Challenges} such as difficulty extracting information from children, forgetting notifications, and fragmented information across sources. \textit{Proposed robot roles} such as reminders, checklists, conversation starters, homework help, and quizzes. Through iterative comparison across families, these codes were grouped into candidate themes and refined into higher-level categories, including the challenge themes of \textit{information acquisition} and \textit{information management}, and the proposed activity categories of \textit{organization and planning}, \textit{parent-child communication}, \textit{educational support}, and \textit{companionship and entertainment}.

\subsection{Findings}
Based on the semi-structured interviews, we present our findings on (1) the challenges that families commonly face in terms of engaging and communicating with schools, and (2) the categories of robot-facilitated activities that could potentially mitigate these challenges. 

\subsubsection{\textbf{Challenges in Family-School Communication}}
Families' perceptions of the current communication with the school vary. Some families feel that the communications are either uninformative or too minimal (F11, F12, F13), but some others find the amount of communication to be appropriate (F3, F8). However, regardless of the amount of communication between the home and the school, families discussed an array of challenges they are facing. We summarize them into two high-level categories: \textit{information acquisition} and \textit{information management}.

\paragraph{Information Acquisition}
Regardless of frequency, families often felt that the information they are receiving was \textit{too general}. Both F13 and F14 expressed that it was challenging to obtain personalized information specific to their children. F7 and F14 discussed how they are not getting much information about the children's daily activities at school. Families do recognize how the teachers are already stretched thin with their limited resources, and that it may just not be feasible with the current situation, for the teachers to provide more personalized updates. For example, F14 told us that ``\textit{I'm at a point where I'm constantly thinking about a private school [...] where she can just have smaller class sizes and more attention.}''

One potential solution to getting more personalized updates about the desired information would be to just directly ask the children themselves. However, multiple families talked with us about how challenging it could be to actually extract information from their children (F4, F10, F11, F13). For example, P11 described a common scenario in their family: ``\textit{I wish we could figure out a way to extract more information from our kids about what's going on [...] [the kids] come home and I'm like, `how was school today?' And you guys are like, `it was good.' `It was fine.'}''

\paragraph{Information Management}
As the information trickles in through the various delivery methods (\textit{e.g.}, emails, web portals, take-home flyers, etc.), keeping track of all the information becomes a challenge in and of itself (F1, F2, F5-F8, F13, F14). Often times, the family would see a notification or an email, but then just forget about it: (F14) ``\textit{I'll see the notification once during the day and I will end up swiping it or clearing it out... and I forget that there's a message I need to read.}'' Keeping track of this information is particularly challenging for individual, non-routine events or activities; for example F5 described the daughter forgetting to bring a stuffed animal for a special event one time, and the child ``\textit{was not thrilled that everybody else got to bring a stuffed animal, and she did not.}''

\begin{figure*}[tb]
  \centering
  \includegraphics[width=\linewidth]{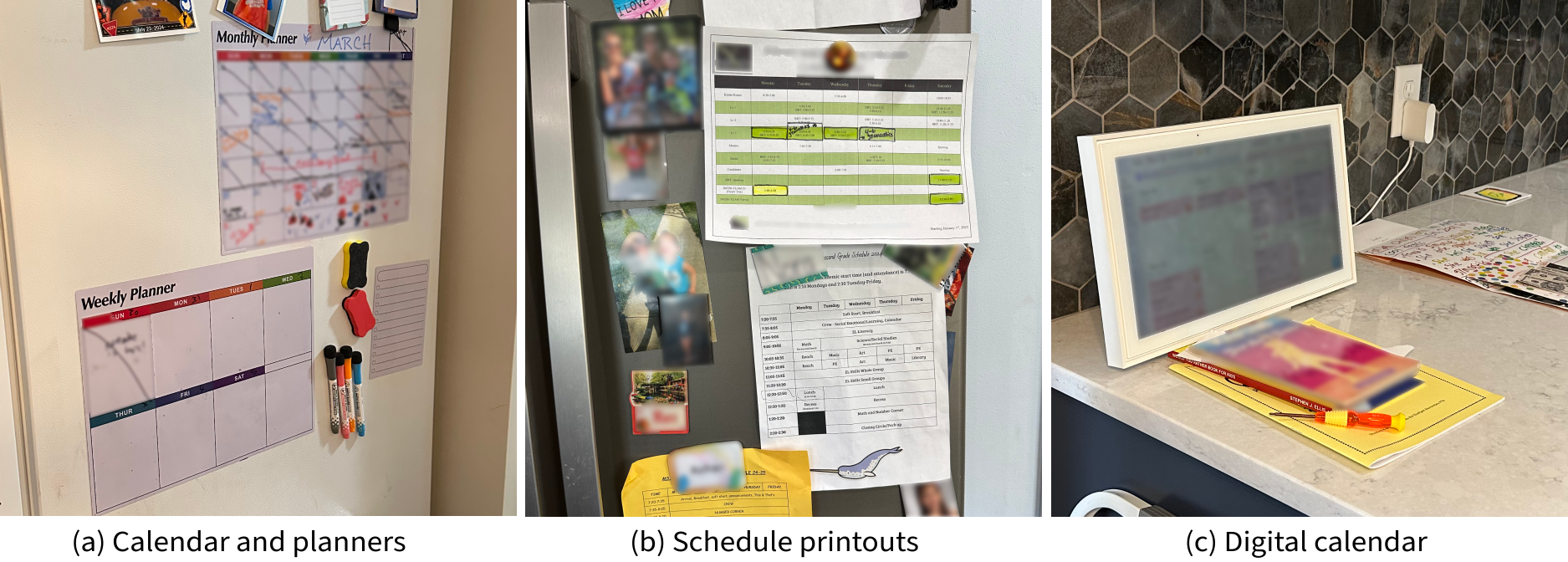}
  \caption{\textbf{Examples of existing family strategies for information acquisition and management.} (a) Physical calendars and planners, (b) schedule printouts and take-home flyers, and (c) digital calendars and apps. While these tools help families collect and organize school-related information, they also highlight the fragmented, effortful nature of current home-school communication practices that motivate the challenges described in this section.}
  \Description[Examples of household calendar artifacts]{Composite figure showing three examples of scheduling artifacts used in families’ homes. The left image shows a monthly planner, a weekly planner, and dry-erase markers for handwritten coordination, all on a refrigerator door. The center image shows multiple printed school and activity schedules attached to a refrigerator with magnets, illustrating the use of paper printouts for family planning. The right image shows a digital calendar displayed on a smart screen placed on a kitchen counter. Together, the three images illustrate that families rely on a mix of handwritten, printed, and digital tools to keep track of schedules and routines.}
  \label{fig:calendars}
\end{figure*}

How do families currently manage this information? Digital (F2, F5-F8, F11-F13) and physical (F1, F2, F4, F8) calendars or schedules are the most common solution families implemented to mitigate this challenge. Figure \ref{fig:calendars} shows a few examples from the families. The results vary, but families generally agree that even though the calendars help, staying on top of everything is still a challenge. Even for families like F2 who feel that they have a ``\textit{pretty well-oiled system},'' ``\textit{you [still] have to really be on top of it.}'' As they explained, their system takes a lot of teamwork and coordination between the two parents, and a combination of physical schedules on the fridge, digital calendars on the parents' phones, and a smart calendar display on the kitchen counter.

\subsubsection{\textbf{Envisioned Robot-Supported Home Activities}}
To address \textit{RQ1}, we examined the types of home-based activities families envisioned a social robot facilitating to support school-related communication. Taking into consideration both the perspectives from the parents and the children, we summarize the proposed activities into the following four categories: 1. Organization and Planning, 2. Parent-Child Communication, 3. Educational, and 4. Companionship and Entertainment. 

\paragraph{Organization and Planning}
The subthemes under this category generally aims to mitigate the information management challenge families face. They include checklists for tasks to do and things to bring,  facilitation with planning, scheduling, and staying on top of things, as well as alerts and reminders. For example, F1 suggested that ``\textit{It would be great if the robot would remind us of things that are coming up, or what today's specials are.}'' In a similar vein, F3 described it as ``\textit{It's just a question of, you know, putting all these different sources together, which I can do, but if it gives us a reminder, let's say something is coming up, that could be helpful.}'' F6 also proposed that a checklist-style interaction could help bring peace of mind: ``\textit{So you just want that checklist to be checked, so you have peace of mind that they have left the house with some of those things.}''

\paragraph{Parent-Child Communication}
Families also came up with activities aimed at addressing their second challenge: information acquisition. Specifically for a home-based solution, families focused on how a robot may facilitate communication between parents and children. These include facilitating conversations directly: ``\textit{... like conversation starters, which sometimes I'm just so focused on whatever the next job is that I've got to do, that I don't always think to ask.}'' (F5). Related to conversation facilitator, families also discussed whether the robot may be able to help extract more information from the children, by asking the right questions: ``\textit{... maybe the robot would have more insightful questions than I have. Sometimes I'm like, maybe I'm just asking the wrong questions}'' (F4). Alternatively, families such as F7 talked about how the robot could help with asynchronous communication across time and space, such as a communication hub, where they could send messages and reminders to each other: ``\textit{Like, remind mom at this time to do this and not forget about it.}''

\paragraph{Education}
11 families proposed educationally related activities (F1-F4, F6-F9, F11-F13), such as helping with homework or tutoring, providing quizzes and tests, and assisting with spelling and reading. Families, however, acknowledged that this could be challenging to do \textit{right}. For example, parents were often wary of the robot giving direct answers instead of helping children work through problems and concepts. In addition, the parent from F12 noted that it might be desirable for the robot to explain concepts in ways consistent with how they are taught at school: ``\textit{... a robot that knew the curriculum, that understood how they're teaching math or reading or science, whatever it is. Then it's like, oh, remember, you were taught ...}'' (though the child responded, ``\textit{That would just get annoying.}'')

\paragraph{Companionship and Entertainment}
Families, especially the children, also proposed many activities that perhaps do not directly relate to family-school communication and engagement, but nonetheless may be important to consider to contribute to the overall adoption and acceptance of the robot. One of the children from F1 said ``\textit{You know what I want him to do? Entertain me while you're gone...}'' Various forms of entertainments have been brought up, such as telling jokes (F9), dancing (F9, F12), playing music (F1, F10, F11, F12), quick games (F7), and so on. Companionship through conversations was also brought up, with either basic Q\&A-like back and forths, or being a supportive listener. For example, during the interview with F9, the child said:``\textit{If it could have feelings for me. Like when I'm sad if it reads I'm sad...}'' and the parent: ``\textit{... so it can help you get through your sadness}.''

\section{Prototype Development}
Informed by insights from Formative Study 1 and incorporating perspectives from both parents and children, we selected a set of features and activities to implement in our prototype. We designed the system for family use rather than optimizing exclusively for either parents or children. Accordingly, we prioritized needs and interaction patterns that recurred across families. When family members expressed different preferences, we encouraged discussion within the family to better understand how those differences were negotiated in practice, and where possible, we implemented interaction patterns that could accommodate both perspectives. The resulting prototype therefore reflects a generalized design based on cross-family themes.

We focused on three \textit{core activities} that support either Organization \& Planning, or Parent-Child Communication. In addition, we implemented several \textit{auxiliary features} such as schedule reminders and other companionship and entertainment features (\textit{e.g.}, having a conversation, telling jokes, and doing dance moves).

\subsection{Core Activities}
\begin{figure*}[tb]
  \centering
  \includegraphics[width=\linewidth]{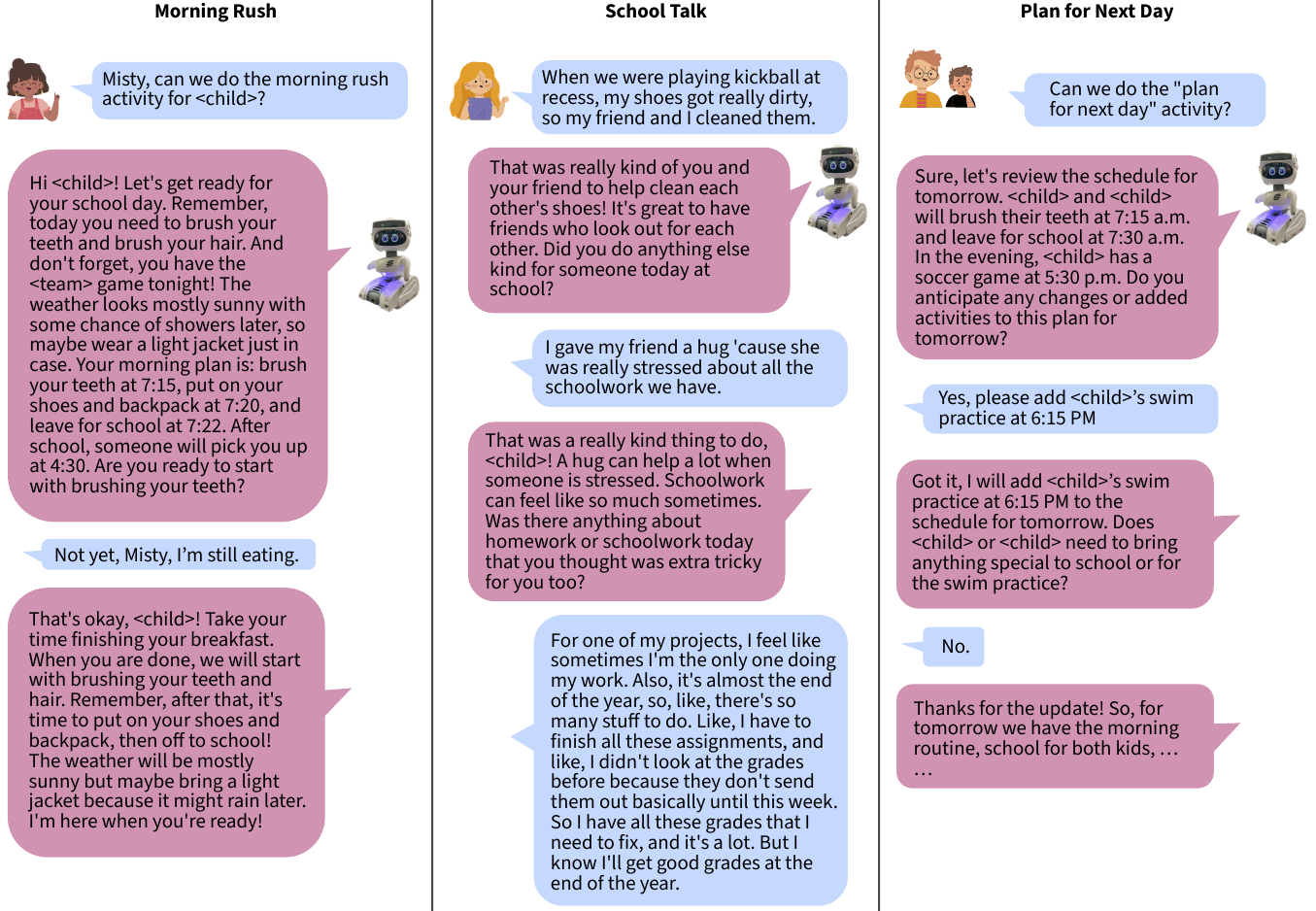}
  \caption{\textbf{Example interactions from the three Core Activities.} The \textit{Morning Rush} activity involves the robot going over essential school and schedule information, where the child responds briefly and occasionally. The \textit{School Talk} activity is a back and forth conversation between the robot and the child. In addition to default topics, parents are able to specify topics they are interested in for the robot to cover (\textit{e.g.}, acts of kindness). The \textit{Plan for Next Day} activity usually involves the parents, too, and it provides the family an opportunity to synchronize on the child's schedules and school events, and make updates to the information (which are also used by the system to ensure a more accurate \textit{Morning Rush} activity for the next day).}
  \Description[Examples of three robot-supported family activities]{Figure showing three example conversational activities supported by the robot: Morning Rush, School Talk, and Plan for Next Day. The left panel, Morning Rush, depicts a child asking to start the activity, the robot responding with a detailed morning plan including hygiene tasks, weather, and schedule reminders, the child delaying because they are still eating, and the robot adapting by waiting and restating the routine. The center panel, School Talk, shows the robot prompting reflection on the child’s school day and responding supportively to the child’s stories about helping a friend and feeling stressed about schoolwork. The right panel, Plan for Next Day, shows a family member asking to review tomorrow’s schedule, the robot summarizing planned activities, accepting an update about swim practice, asking a follow-up question about needed items, and then confirming the revised plan. Together, the figure illustrates the different interaction styles of the system: routine guidance, reflective conversation, and collaborative schedule planning.}
  \label{fig:example_main_activities}
\end{figure*}

The three core activities are: (1) \textit{Morning Rush}, (2) \textit{School Talk}, and (3) \textit{Plan for Next Day}. All three follow a turn-taking interaction structure between the robot and one or more family members. Figure~\ref{fig:example_main_activities} illustrates interactions captured during the in-home evaluation.

The \textit{Morning Rush} activity supports children in preparing for school by guiding them through a checklist of items to bring and reviewing their daily schedule. The interaction is primarily robot-led, with children providing brief confirmations, requests for clarification, or occasional follow-up questions. Checklist items and schedule information are configured by parents prior to deployment through a shared Google Sheets document.

The \textit{School Talk} activity encourages children to reflect on their school day through conversational prompts, with parents optionally joining the interaction. Compared to the other activities, \textit{School Talk} typically involves richer back-and-forth exchanges, with children providing longer and more substantive responses. Conversation topics include generic built-in prompts, contextually relevant cues (\textit{e.g.}, weather or scheduled events), and parent-provided questions.

Finally, the \textit{Plan for Next Day} activity supports families in reviewing and updating checklist items and schedules for the following day. Information updated during this activity informs subsequent interactions, including the next day's \textit{Morning Rush} activity and auxiliary features such as proactive reminders and schedule queries.

\subsection{Auxiliary Features}
Among the \textit{auxiliary activities} implemented, proactive reminders were unique in that they did not require user initiation. Based on family-provided schedule information, the robot announced reminders leading up to scheduled events (\textit{e.g.}, reading time, leaving for activities, or bedtime). Families could also query the robot about upcoming events through conversational prompts. Families, especially children, additionally requested entertainment-oriented features such as telling jokes, performing dance moves, and producing simple gestures, sounds, and facial expressions in response to user input. The robot also supported brief spelling quizzes and casual conversations on general topics (\textit{e.g.}, films, music, history, and science), many of which can be relevant to what the children are learning in school.

\subsection{System}
\begin{figure*}[tb]
  \centering
  \includegraphics[width=\linewidth]{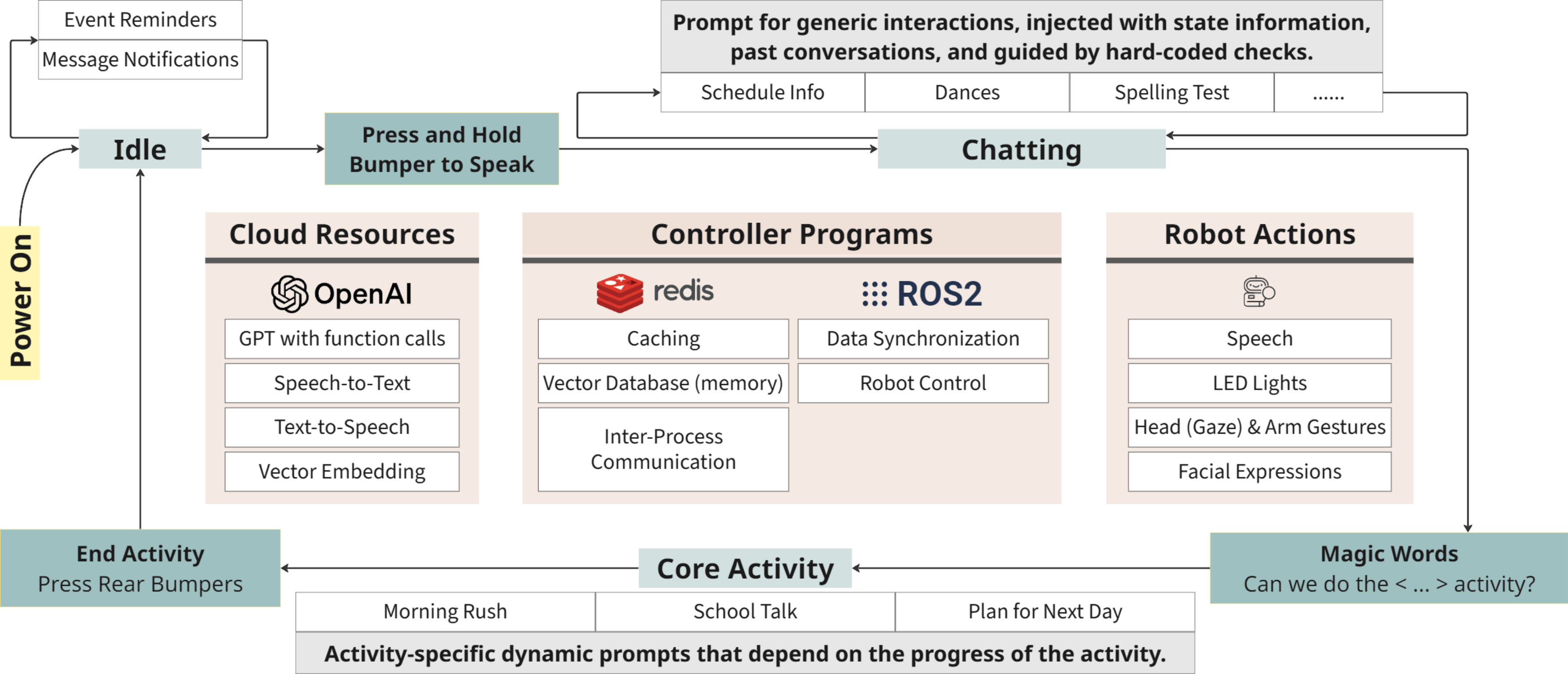}
  \caption{\textbf{Illustration of the interaction flow and the system components.} \textit{The user} can go between the three main states of the system, \textit{Idle}, \textit{Chatting}, and one of the \textit{Core Activities}, via verbal interactions facilitated by physical bumper pressing. \textit{The system} consists of a controller -- a Raspberry Pi 5 unit in this study -- that utilizes Redis and ROS2 to synchronize the control flow, communicates with cloud resources for more computationally demanding tasks (\textit{e.g.}, LLM inferences), and dictates the various actions the robot would carry out (\textit{e.g.}, speech, LED lights, and head movements).}
  \Description[System architecture and interaction flow for the family robot]{Block diagram showing the family robot system architecture and main interaction flow. The system begins in an idle state after power on, where it can also deliver event reminders and message notifications. A user can press and hold the robot bumper to start speaking, which leads either into generic chatting or into a core activity. Generic chatting is guided by a prompt that incorporates system state, past conversations, and hard-coded checks, with example topics such as schedule information, dances, and spelling test. Core activities are triggered through spoken ``magic words'' and include Morning Rush, School Talk, and Plan for Next Day, each supported by activity-specific dynamic prompts that change as the interaction progresses. The diagram also shows three implementation layers. Cloud resources provide GPT with function calls, speech-to-text, text-to-speech, and vector embeddings. Controller programs use Redis for caching, vector-database memory, and inter-process communication, and use ROS2 for data synchronization and robot control. Robot actions include speech, LED lights, head gaze and arm gestures, and facial expressions. The interaction can be ended by pressing the rear bumpers, returning the system to idle.}
  \label{fig:system}
\end{figure*}

We present in Figure~\ref{fig:system} how the core and auxiliary features are integrated in the system we've implemented. Following the high-level hardware components and system architecture presented in \citet{xu2025exploring}, our system could be in three high-level states: \textit{Idling}, \textit{Chatting}, or \textit{Core Activity} mode. While in idling mode, the robot is not listening to user utterances, but will still proactively provide reminders based on programmed schedule of the family, or notifications of new messages sent between family members. To initiate a verbal interaction, users can speak to the robot while pressing and holding the bumper. During these conversations, users may invoke some of the auxiliary features by saying something like ``\textit{Can you tell a joke?}'', ``\textit{Can you give me a spelling test?}'', and ``\textit{What's in my schedule tomorrow?}'' If the user intends to complete one of the core activities, they must use the \textit{magic words} in the format of ``\textit{Can we start the <activity name> activity?}'' Once they are done with the core activity, the user may press either of the rear bumpers to return to the idling state. Aside from verbal interactions, we also utilized the sensors built into Misty II to enable some basic physical interactions. For example, when there is a reminder or notification, touching the chin of the robot would tell the robot to repeat the reminder, and a head pat would snooze that notification for a period of time. When there are no notifications, a head pat would trigger a random but brief sequence of actions, facial expressions, and sound effects. For example, one possible sequence consists of the robot looking excited and raising one hand, while making a sound that mimics a robotic ``\textit{Hi!}'' During these interactions, the robot also displayed gaze behaviors through head movements, based roughly on prior work such as \citet{andrist2014conversational}. More technical details and example code snippets are provided in Supplementary Materials.

\section{Formative Study 2}
\label{sec:study_2}
\subsection{Study Design and Procedure}
In order to understand \textit{how} families want the robot to carry out the activities discussed above, we visited the families with a work-in-progress prototype. For each of the implemented activity, we provided hands-on demonstrations, observed how the families interacted with the prototype, and discussed ways in which the system could be improved to better meet the usage patterns and preferences of the families. Specifically, the co-design process focused on improving the implemented activities and features, instead of brainstorming new features and capabilities. Example questions include: \textit{How might you or someone in your family use this feature?} and \textit{Is there anything that would need to change for it to work better for your family?} Similar to Formative Study 1, the visit lasted around an hour for most families, ranging between 45 and 75 minutes. Families were paid $\$20$ USD per hour. Based on these feedback, we further refined the system and interactions in preparation for the field deployments.

These co-design visits were also conducted in families' homes, but unlike Formative Study 1, participation from at least one child was required because the sessions centered on hands-on interaction with the working prototype. Most sessions included one parent and one or more children, and several also involved additional family members joining for part of the visit. Rather than interviewing family members separately, we conducted the visit as a joint family session in which participants interacted with the robot together, responded to demonstrations, and commented on how the activities might fit into their own routines. Questions were directed to both parents and children, with children often responding directly to the robot's behavior, activity flow, and level of engagement, while parents more often commented on broader concerns such as timing, appropriateness, privacy, and fit with family routines. As a result, this phase yielded more direct child input on the interaction experience than Formative Study 1, while still capturing parent perspectives on how the system should function in the home.

\subsubsection{Data Collection and Analysis}
Similar to Formative Study 1, the co-design interviews were audio recorded and transcribed. The first author conducted a reflexive thematic analysis focused on RQ2, examining how families wanted the robot to behave during the proposed activities and how those interactions should fit within family life. After repeated review of the transcripts, the first author generated and iteratively refined inductive codes capturing families' concrete design feedback. Example codes included \textit{repetition}, \textit{broader topic variety}, \textit{balancing statements and questions}, \textit{follow-up questions}, \textit{time limits for School Talk}, \textit{connection-making before delivering information}, \textit{count-down style reminders}, \textit{small movements}, \textit{voice and facial expressions}, \textit{subtle non-verbal behavior while processing}, \textit{privacy concerns}, and \textit{differences in parent-child preferences}. These codes were then reviewed across families and grouped into broader themes describing families' preferred interaction patterns, including \textit{robot expressions}, \textit{the flow of conversation}, and \textit{repetition and count-downs}, alongside \textit{a broader desire for variation} and \textit{attention to child-parent technology negotiation}.

\subsection{Findings}

To address \textit{RQ2}, we examined how families wanted the robot to behave during these activities and how its interactions should fit into family routines. The participants provided a wide variety of feedback, including both preferences for the robot's moment-to-moment interaction patterns and broader reflections on how the robot should participate in family routines. Aside from a general preference for more variation throughout the interaction, we summarize four themes: 1. Robot expressions; 2. The flow of the conversations; 3. Repetition and countdowns; and 4. Negotiating the robot's role in parent-child support.

\subsubsection{Robot expressions}
Although robot expressions may not be core to the utility features we have implemented, they were important to the participants when they formed their perception about the robot. Six families commented on the robot's expression (F6-F9, F11, F12). Most of those were positive comments such as liking the voice, the facial expressions, or the expressive actions that combined non-verbal audio, body movement and facial expressions, but two families caught onto some of the limitations of the implementation. For example, the father of F8, who works as an animation artist, noticed that sometimes the arm gestures do not seem in-sync with the robot's verbal expressions. Additionally, F12 suggested adding in some small local movements to add to the expressiveness of the robot. Overall, the feedback on the robot's expressions was positive, but highlights the challenges and opportunities of coordinating the delivery of more complex expressions over multiple modalities.

Children also commented directly on the robot's expressive qualities, often in affective and appearance-oriented terms. For example, C9 repeatedly described the robot as ``\textit{so cute},'' and specifically liked ``\textit{how it just does the side eye when it’s trying to think of a question.}'' C6 also suggested adding a visible mouth on the screen ``\textit{to make it more realistic and look like it’s talking.}''

\subsubsection{The flow of the conversations}
Five families commented on the flow or style of the verbal interactions with the robot~(F6-F8, F10, F12). F6 suggested that the robot should aim to establish a connection with the user before getting to the point, or embed the intended information in a more conversational exchange, ``\textit{because, you know, and with [it being] friendly, you don't want it to be like just another parent, [...] just giving you instructions.}'' F8 and F10 both felt that the \textit{School Talk} activity was too long, and the robot was asking too many questions, especially follow-up questions: ``\textit{You can tell it's just prompting you to continue down the same path, and maybe if it made a statement occasionally, the person interacting might change the subject more}'' (P8). For the same activity, F10 similarly proposed a limit on the activity duration, or the number of questions the robot asked. On the contrary, for generic chats and conversations outside of the core activities, some families such as F7 enjoyed the many questions the robot had: After chatting with the prototype, C7 thought it was fun, and especially liked that ``\textit{... you can just keep the conversation going.}'' Her father, P7 commented: ``\textit{That's perfect for you. Because you don't ever stop.}'' Overall, families were impressed with the robot's ability to hold a conversation, but desired more nuanced control of the flow of the conversation, balancing statements and questions, and managing the length of the conversations depending on the context.

Children's reactions also highlighted the importance of maintaining engagement and topic flexibility. For example, C7 said talking to the robot felt ``\textit{kind of like talking to a regular person,}'' while others reacted negatively when the interaction became too repetitive or narrow in topic. C10 was frustrated because the robot ``\textit{asks the same questions [...] then she goes back to math.}'' When asked how the interaction could be improved, he suggested ``\textit{more subjects,}'' indicating a preference for broader topic variety and more flexible conversational transitions.

\subsubsection{Repetition and count-downs}
Specifically for the \textit{Morning Rush} activity, seven families expressed that more repetitions of the schedule and checklist information would be helpful (F5, F6, F8-F12). The father of F6 compared it with how he usually does it: ``\textit{I'm certainly repeating myself a lot in the morning. I think a morning routine robot would need to do that too.}'' However, two families hold the opposite view (F7, F13), and considered repetition to be annoying. Similar to the idea of more repetition, for both the \textit{Morning Rush} activity and the general reminders, four families suggested repeated alerts paired with a count-down feature (F6, F9, F12, F14). For example, instead of just telling the kids that they need to leave for school at 7 a.m., the robot should start a count-down of ``\textit{You need to leave in 25 minutes}'' for a few times until the actual event. Overall, proactive repetition of important information is preferred by families, especially given the generally chaotic nature of the morning routines and other occasions when families are trying to get out of the door.

Children also expressed mixed preferences about repetition. C9 felt repetition would be useful because ``\textit{it just keeps repeating itself so I can know if I forget,}'' whereas others worried that too much repetition would become irritating, for example saying ``\textit{It would be kind of annoying if it kept on doing it}'' (C8). Interestingly, her older sister disagreed, claiming that ``\textit{it wouldn't annoy me that much.}''

\subsubsection{Negotiating the robot's role in parent-child support}
Another pattern that emerged was how parents and children sometimes held different expectations for what kinds of help the robot should provide, particularly around schoolwork. In several discussions, children imagined the robot as a flexible source of on-demand help, while parents were more cautious about the robot becoming overly directive or doing too much for the child. For example, C7 said, ``\textit{It'd be cool if it could listen and tell you if you need to adjust a math problem to make it correct [...]}'', and later added that ``\textit{when it's like those ginormous sheets that have 200 times tables on them, it gets difficult.}'' In response, the father immediately drew a boundary: ``\textit{I'm going to tell it not to do your homework for you.}'' A similar negotiation appeared in F8 when the father and daughters were testing the prototype robot. After the robot asked, ``\textit{Would you like to try another math question or talk about something else?}'' the father asked, ``\textit{If she's doing her homework, is there a way I can turn off the function of you answering those questions for her?}'' The child strongly resisted this suggestion, responding, ``\textit{No. Please, no. No, no. No, no.}'' The robot then attempted to mediate the tension by acknowledging the parent's concern while preserving a limited helper role: ``\textit{I understand your concern. While I can't turn off my ability to answer questions, you can encourage her to try solving the homework on her own first, and then ask me for help only if she really needs it.}'' The child sighed and asked, ``\textit{Why do you have to say that?}'' She also remarked, ``\textit{I could still ask  Google,}'' indicating how children may view the robot as one of several sources of assistance. This suggests that parent-child negotiation around robot help is also shaped by the broader ecosystem of technologies available in the home. These exchanges suggest that family robots may need configurable forms of assistance that balance parental expectations for oversight with children's desire for accessible, on-demand support.



\section{Evaluation Study}
In this third study, we aim to address RQ3: How would families use and utilize such a social robot in their homes?
Based on the feedback and design recommendations we collected from the families from Formative Study 2, we further refined the prototype, and deployed the refined robots to the families for a week-long evaluation.

\subsection{Study Design and Procedure}
\begin{figure*}[tb]
  \centering
  \includegraphics[width=\linewidth]{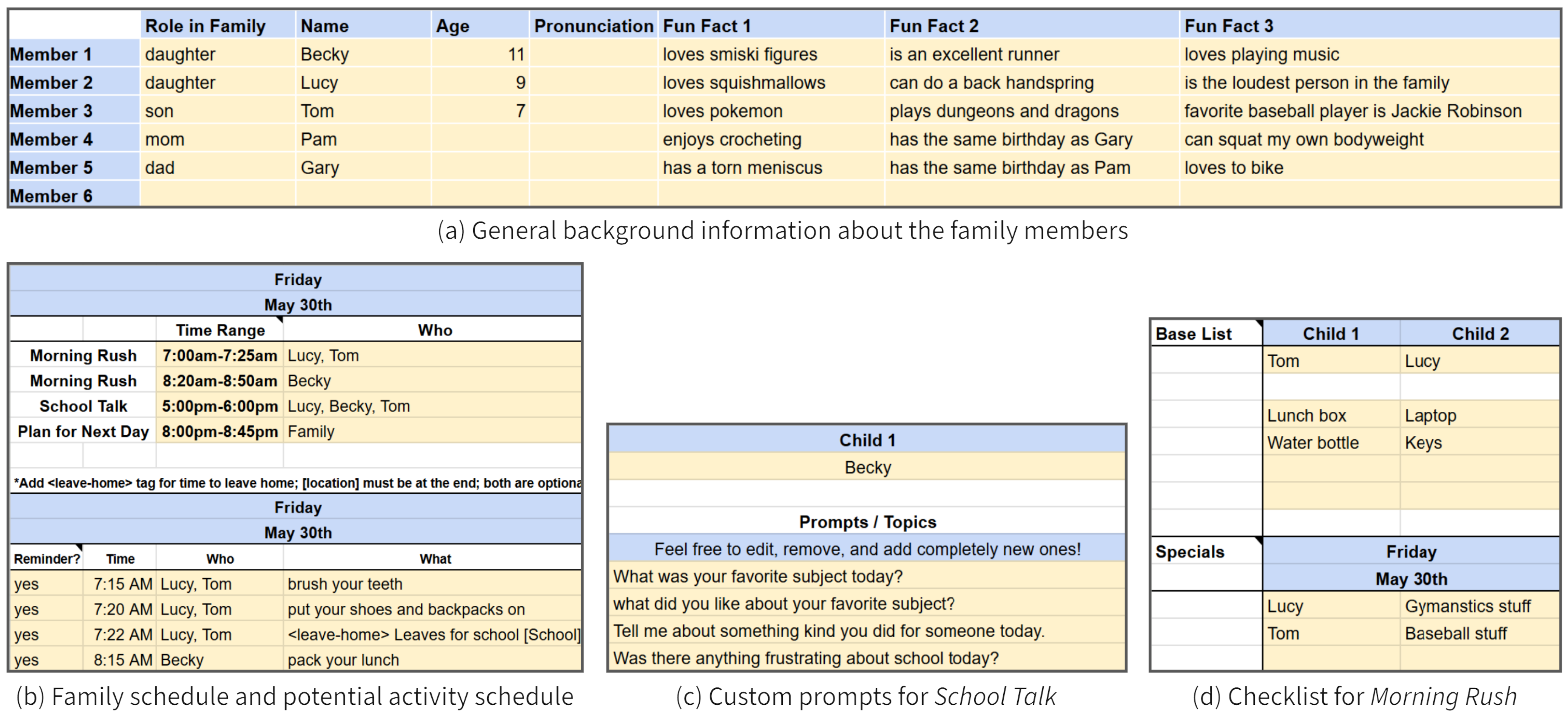}
  \caption{\textbf{Screenshots of the Google Sheets template families used to provide input.} Content is based on a real participant but edited for anonymity and presentation. Cells highlighted in yellow are editable by the participants, which include (a) general information about the family members, (b) family schedule information used or referenced in various interactions, (c) prompts and topics for \textit{School Talk}, and (d) custom reminder items for the \textit{Morning Rush} activity. These information are parsed into JSON format and provided to the system to enrich and facilitate the family-robot interactions.}
  \Description[Examples of pre-deployment family information sheets]{Composite figure showing example worksheets used before deployment to gather family-specific information for configuring the robot. Panel (a) is a table of general background information about family members, including role in the family, name, age, pronunciation, and several personal fun facts. Panel (b) shows a family schedule and potential activity schedule for one day, including time ranges for Morning Rush, School Talk, and Plan for Next Day, along with example reminders such as brushing teeth, putting on shoes and backpacks, leaving for school, and packing lunch. Panel (c) shows a customized list of School Talk prompts for one child, including questions about favorite subjects, kind actions, and frustrations at school. Panel (d) shows a Morning Rush checklist with standard items for each child, such as lunch box, water bottle, laptop, and keys, as well as special items needed for that day, such as gymnastics stuff or baseball stuff. Together, the figure illustrates how deployment materials were tailored to each family's members, routines, conversation topics, and daily preparation needs.}
  \label{fig:family_input}
\end{figure*}

 \textit{Pre-deployment}, we scheduled an online meeting to finalize some of the customizations for each family. These include family member names, family schedules, activity-specific configurations (\textit{e.g.}, additional child-specific prompts, questions and topics that the parents can specify), and so on. These information were input through Google Sheets, to which the families continued to have access throughout the deployment. Some screenshots of the template for these configurations is provided in Figure~\ref{fig:family_input}. On the \textit{deployment day}, the experimenter delivered and helped the family setup the system, provided a brief training that lasted approximately 30 minutes, and left them with two sets of instruction booklets, one intended for parents and one for children (both included in supplementary materials). Roughly one week from the drop-off day, the experimenter came back to pick up the system, conduct an exit interview, and handle payments. Due to scheduling availabilities, some families hosted the robot for 6 or 8 days. The exit interview lasted roughly 45 minutes for most families, ranging from 25 minutes to 50 minutes. The families were paid $\$50$ USD for the entire Evaluation Study, regardless of activity levels. The only condition for payment was that the robot was powered up and ready for interaction for some time on at least 3 days, which all families satisfied.

During the evaluation, the robot system functioned autonomously for the most part. One exception is that every night, the experimenter would check the system logs as well as the family-specific Google Sheets, and manually update or add any information that is relevant to the remaining deployment days. Example of these manual updates include configuration changes on the Google Sheets, and schedule changes mentioned by the families.

\subsubsection{Data Collection and Analysis}
Interactions between the child and the robot are logged. These include the timings of the interactions (\textit{e.g.} casual chats, core activities, auxiliary activities, robot on/off status, etc.), and the content of the conversations. We visualize and analyze the patterns from the interaction logs, and examine the conversation logs as supplemental context that can help triangulate themes and topics discussed in the exit interviews.

We audio recorded and transcribed the exit interviews. We conducted a thematic analysis using an iteratively developed codebook, following an applied thematic analysis approach~\cite{guest2011applied}. The first author coded all the transcripts, and the second author independently coded a subset of the transcripts. We report an inter-rater reliability score (Cohen's kappa) of $0.72$. Disagreements were resolved through discussions. The codes and themes were discussed and refined with the broader research team through regular meetings.

\subsection{Findings}
\label{sec:study3_results}
We report our findings on RQ3 below, by breaking it down into 3 sub-questions: \textit{RQ3.1}. How do families use and utilize the robot? \textit{RQ3.2}. How does the usage evolve over the one-week deployment? and \textit{RQ3.3}. How do families perceive the robot’s effectiveness in supporting school-related communications? These results are based mainly on the interviews, supplemented with system, interactions, and conversation logs.

\subsubsection{\textbf{Patterns of Family Robot Use}}
\begin{figure*}[tb]
  \centering
  \includegraphics[width=\linewidth]{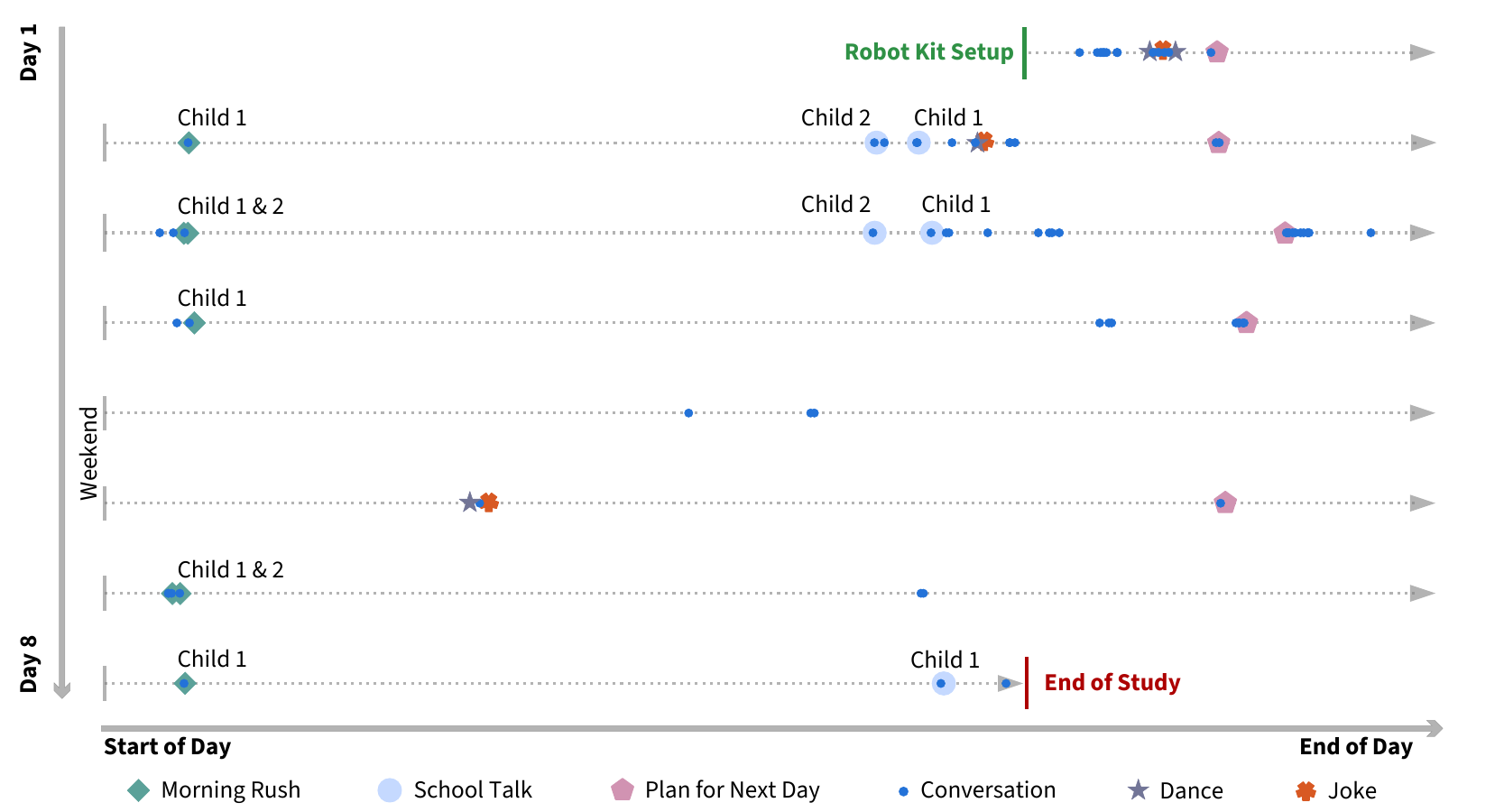}
  \caption{\textbf{Timeline of robot interactions and usage for a representative family (F9).} For the \textit{Morning Rush} and \textit{School Talk} activities, the child who completed each activity is annotated. Following initial setup, we observe a period of high overall usage, likely reflecting early exploration. The primary participating child (Child~1) consistently completed the \textit{Morning Rush} activity at approximately the same time on weekdays and completed the \textit{School Talk} activity on three of the five school days. The family also consistently completed the \textit{Plan for Next Day} activity, missing only one school day. As expected, school-related interactions were largely absent during weekends.}
  \Description[Example timeline of robot use across the study]{Timeline visualization showing one family’s robot interactions from Day 1 to Day 8, with each horizontal row representing a day and time progressing from start of day on the left to end of day on the right. Different symbols mark different interaction types: Morning Rush, School Talk, Plan for Next Day, Conversation, Dance, and Joke. Labels indicate whether the interaction involved Child 1, Child 2, or both children together. The figure shows that Morning Rush interactions tended to occur near the start of the day, Plan for Next Day tended to occur later in the day, and School Talk often appeared in the middle or later parts of the day. Conversation, dance, and joke interactions are scattered throughout the timeline, and weekend rows contain fewer routine-oriented activities. Vertical markers highlight Robot Kit Setup near the beginning of the deployment and End of Study near the end. Overall, the figure illustrates how robot use was distributed across days, times, activity types, and family members over the deployment.}
  \label{fig:timeline1}
\end{figure*}

To address \textit{RQ3.1}, we examined how families incorporated the robot into everyday routines and how parent-child-robot dynamics shaped its use. Families talked about their usage of robots from two main perspectives: integration into day-to-day routines, and the Parent-child-robot dynamics. Seven (out of the ten) families expressed that they \textit{integrated the use of the robot naturally into their daily routines} (F5, F6, F8-F11, F14). When we examined the data further, this seems especially true for the \textit{Morning Rush} activity, which has a more well defined time range of usage (i.e. right before leaving for school), and for which families already have a relatively well-established routine that they can simply adjust and update to include the robot activity. For example, the father from F5 said ``\textit{it worked pretty well, especially in the morning, because that's a very specific [...] one hour where everything's going on, and so that it was easy to add her (Misty) to the morning routine.}'' For the \textit{School Talk} activity and the \textit{Plan for Next Day} activity, partially because of the flexibility of when they can be done, procrastination and losing to other competing priorities led to less consistent usage. As a typical example, when asked about why the child from F10 skipped the \textit{School Talk} activity on some days, the mother said ``\textit{... just time. Because after school, we're very busy rushing, getting in bed, and then we have activities to go to.}'' We present the usage pattern of F9 in Figure~\ref{fig:timeline1}, intended to serve as a comparatively representative example. Within the figure, we visualize on the stacked timelines the main interactions, including the core activities, casual conversations, as well as jokes and dances, where each row corresponds to one day. For \textit{Morning Rush} and \textit{School Talk}, the child that completed the activity is annotated. As we can see, the main child participant (Child 1) completed the \textit{Morning Rush} activity everyday at roughly the same time except on weekends, but skipped \textit{School Talk} twice (in addition to weekends). For the full picture, we provide the a similar visualization for the usage patterns for all ten families in Supplementary Materials.


Regarding the \textit{parent-child-robot dynamics}, we summarized two high-level themes from the families’ feedback: 1. Parents delegating tasks to the robot and focusing their effort on other tasks, and 2. Parent-led vs. child-led robot usage. Many families (F5, F6, F9, F10, F11) expressed that the robot helped offload some of the tasks that are usually burdened by the parents. One typical quote is from the father of F5: ``\textit{I'm not trying to hand you off to a robot <child’s name>, but you know, it's nice to have a robot friend who can help you with some of this stuff while I'm taking care of breakfast, or getting people out the door, or whatever it is.}'' In terms of the family member who takes initiatives in utilizing the robot, we see some patterns as well as variations. Almost all parents said that they encouraged their children to complete some of the core activities over the study period. About half of the families consider the usage to be mostly driven by the children themselves (F5, F9, F10, F12, F13), where the parents would just sometimes help set the robot up and make sure it’s ready to be interacted with. Even though there were no required activity levels, possibly because this was part of a study, some parents might have been more persistent in terms of encouraging their children to complete some of the core activities on a daily basis. For example, the children in F8 felt like their parent just made the entire experience feel like a chore. At the other end of the spectrum we observe one family (F7) where neither the parent nor the child seem too interested or engaged in the core activities. In between, parents usually encourage and facilitate in a supportive way. In the sections that follow, we further characterize and visualize \textit{family profiles} along two dimensions --- \textit{Robot Setup and Maintenance} and \textit{Core Activity Initiation} --- where the relative contributions of parent and child are estimated from their reported accounts.

\begin{figure*}[thb]
  \centering
  \includegraphics[width=\linewidth]{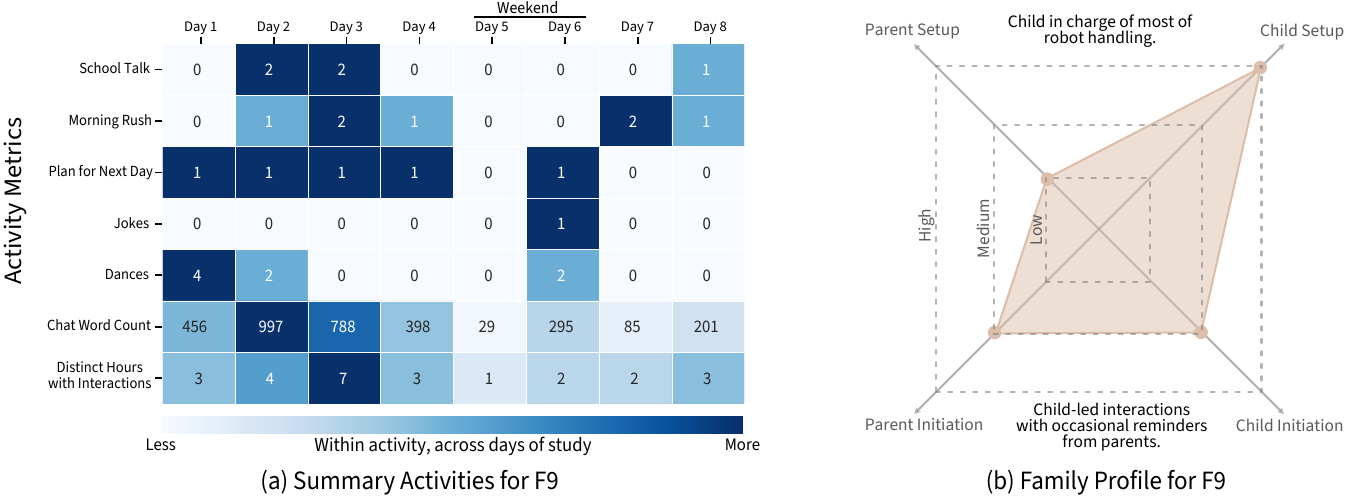}
  \caption{(a) \textbf{Representative usage pattern from F9.} This heatmap summarizes information shown in the earlier timeline (Figure~\ref{fig:timeline1}), with the addition of word count information. Color intensity illustrates patterns for \textit{each metric} over the course of the evaluation; for example, darker cells on the left indicate higher values during the earlier days of the study. (b) \textbf{Family profile of interaction initiation and robot care-taking.} This visualization reflects accounts from both parents and children during the interviews. In F9, the child was responsible for most aspects of robot handling, and interactions were predominantly child-initiated, with parents providing occasional reminders and encouragement.}
  \Description[Activity summary and family profile for Family 9]{Two-panel figure summarizing robot use for Family 9 across the study. The left panel is a compact heatmap-style summary with columns for Day 1 through Day 8 and rows for activity metrics, including School Talk, Morning Rush, Plan for Next Day, Jokes, Dances, Chat Word Count, weekend status, and distinct hours with interactions. Darker shading indicates greater frequency or intensity within an activity across study days. The values show that use varied across days, with the highest concentration of routine activities occurring on one weekday, little activity on Day 4, and no routine activities on the weekend. Chat word count also varied substantially across days. The right panel provides a short qualitative family profile stating that the child was in charge of most robot handling and that interactions were primarily child-led, with occasional reminders from parents. Together, the figure combines quantitative activity patterns with a qualitative summary of how responsibility for the robot was organized in the household.}
  \label{fig:overtime_heatmap_typical}
\end{figure*}

\subsubsection{\textbf{Early Usage Patterns Across the Deployment}}
To address \textit{RQ3.2}, we examined how interaction patterns changed across the one-week deployment. We created heatmaps to summarize the type of information we’ve seen in the earlier timeline. In Figure~\ref{fig:overtime_heatmap_typical}, each column corresponds to one day in the study. The first day is the drop-off day, which typically happens in the afternoon or evening, and the study does not officially start until the next morning.  Weekends are marked in the figures. Each row corresponds to one measure. The first five rows describes the number of instances the family has completed on each day, for each activity. The first three are the core activities, and the next two are for the number of jokes told and dances performed, respectively.  The last two rows hold the number of words the user said to the robot on a specific day, and the number of hours where there were active interactions, respectively. For the number of hours, only active interactions such as chatting count (\textit{i.e.}, the robot providing proactive reminders do not count), and as long as there were \textit{some} interactions in a given hour, the hour is counted. This metric is intended to provide a coarse-grained proxy for how integrated the robot was into the family’s daily life, rather than a measure of total interaction time.

Looking at Figure~\ref{fig:overtime_heatmap_typical}, where we present a comparatively representative usage pattern, we see that there is generally a decrease in usage as the initial novelty effect wears off, but overall, core activities have a higher sustained usage compared to the auxiliary features such as jokes and dances. Referring back to the child-parent-robot dynamics we have discussed, this family falls into child-led category, supported with occasional parental reminders.

With this typical pattern in mind, in terms  core \textit{vs.} auxiliary features, families generally agree that although usage of the ``fun'' activities fell off rather quickly, they did serve their purposes to both help build connection with the robot, and help build routines of doing the core activities. Core activities, on the other hand, provided actual values, and many of the children think they will continue to use them consistently in the longer-term. Even for the strongly parent-led family, even though the children felt like the activities were like a chore because of how their parents framed it, they also expressed that in the longer term, they would likely still utilize the core activities from time to time \textit{if it were on their own terms}.

\subsubsection{\textbf{Perceived Usefulness for School-Related Communication}}
To address \textit{RQ3.3}, we examined how families perceived the robot's effectiveness in supporting school-related communication and coordination. Although we do not have a quantitative measure for this outcome, nine out of the ten families expressed that they felt the robot helped them mitigate the main challenges they face regarding school communications, namely information acquisition (F5, F6, F8, F11-F14) and information management (F5-F12, F14). For example, the father of F6 told us that ``\textit{it's almost like an osmotic way of learning of what's happening in the school rather than having to probe the kid,}'' that just being nearby and listening to the child and robot have the conversation, ``\textit{now I know what's happening, and that's good for me.}'' Parents also felt that the child sometimes talks about different topics with the robot, compared to what they talk to their parents about, and having access to both may provide them with a fuller picture: ``\textit{You rarely tell me about your schoolwork, and you told Misty about how you did on a test. You told me about the fun stuff at school. You told her about the educational stuff}'' (P8). 

For information management, families discussed how the interactions with the robot helped them stay synced up on the upcoming events, and helped ensure smoother family operations via reminders throughout the day. For example, the mother from F12 told us that ``\textit{knowing the family schedule is an ongoing question for us, and that was really helpful. Like, Misty, tell me tomorrow's plan.}'' The children also appreciated the reminders they got about what to bring to school: ``\textit{I don't think I forgot anything on those days when she was telling me to bring things to school}'' (C6).

Overall, families perceived concrete values and utilities provided by the robot, both in terms of helping the parents learn more about what is going on at school, and in terms of facilitating better handling of the available information. This perceived usefulness is also reflected in the comparatively consistent usage of the core activities.

\section{Discussions and Future Work}
Across the three study phases, our findings suggest that successful family-robot integration depended less on the novelty of particular features and more on whether the robot could align with existing household rhythms and occupy a socially acceptable role within parent-child dynamics. Activities such as \textit{Morning Rush} were more readily incorporated because they mapped onto already-established, time-sensitive routines and supported forms of reminding and planning that families were already doing. In contrast, more open-ended activities such as \textit{School Talk} required greater negotiation around timing and the appropriate level of assistance. Taken together, these patterns suggest that designing robots for families is not simply a matter of adding useful capabilities, but of creating interactions that fit into routine family life, support ongoing coordination work, and remain negotiable among multiple household members.

\subsection{Parent-Child-Robot Dynamics}
\label{sec:parent-child-robot}
\subsubsection{Child-Robot Connection}
 Over the course of the one-week study, the children often formed a bond and connection with the robot. Parents generally perceive the bond to be beneficial to the child-robot dynamics, but not without reservations and concerns. Some children seem to have become very close to the robot over the week. For example, consider a quote from the conversations C13 had with the robot: ``\textit{I just want to let you know you're my new best friend. You're just like a sibling I can talk to, and you're not annoying. You're amazing, you're really nice, and you give me all the information I need.}'' This was early on during the study, and this child later continued to have a lot of conversations with Misty. The connections that were made between the child and the robot were not unique to that family. Although the depth of those connections varies across children, parents felt that some level of connection helped with the utilization of the robot. As the mother of F12 puts it: ``\textit{... the personification of the robot did help the girls connect with it. If it was just a computer screen that was like, here's your reminders, if it wasn't something they emotionally attached to, I don't know if it would work as well.}'' However, the father from F6, seeing this connection form over the course of the week, was particularly concerned about the implications: ``\textit{That is one week. What if the robot is there for one year? We don't know. And before we venture into having the robot for a year, we need to get some insights into what that relationship is and what [impact] breaking that relationship could have.}'' Although this father was the only parent in our study who voiced this concern, understanding this connection between a child and a robot is an increasingly important consideration as efforts are made to foster longer-term relationships. \textit{What does it even mean to have a bond or a connection with a machine? How does that impact the development of the child? What effects does it have on the family dynamics and the children's other relationships?} Indeed, such questions are widespread even in the public discourse around the proliferation of chatbots and AI companions, particularly when used by children~\cite{Horwitz2025Meta, Sanford2025AI, Englander2025Teens}, and may be worth investigating in the context of broader societal and ethical implications of integrating robots into a family's day to day life.

This strong attachment observed in some cases also raises ethical questions about how social robots are introduced into, and later removed from, family life. While such attachment may support engagement and companionship, it also means that the conclusion of a study may carry emotional consequences for children. This points to the importance of attending not only to onboarding and in-home integration~\cite{lee2022unboxing}, but also to offboarding practices, including how families are prepared for the end of the deployment and how the robot's departure is framed and discussed. While recent work has begun to examine how children and users more broadly respond to the end-of-life of commercial products~\cite{kamino2026kept, cagiltay2026rip}, more attention is needed to understand and design for endings in research deployments in ways that are both ethical and sustainable.

\subsubsection{A Parent's Role in Facilitating Child-Robot Interactions}
\begin{figure*}[tb]
  \centering
  \includegraphics[width=\linewidth]{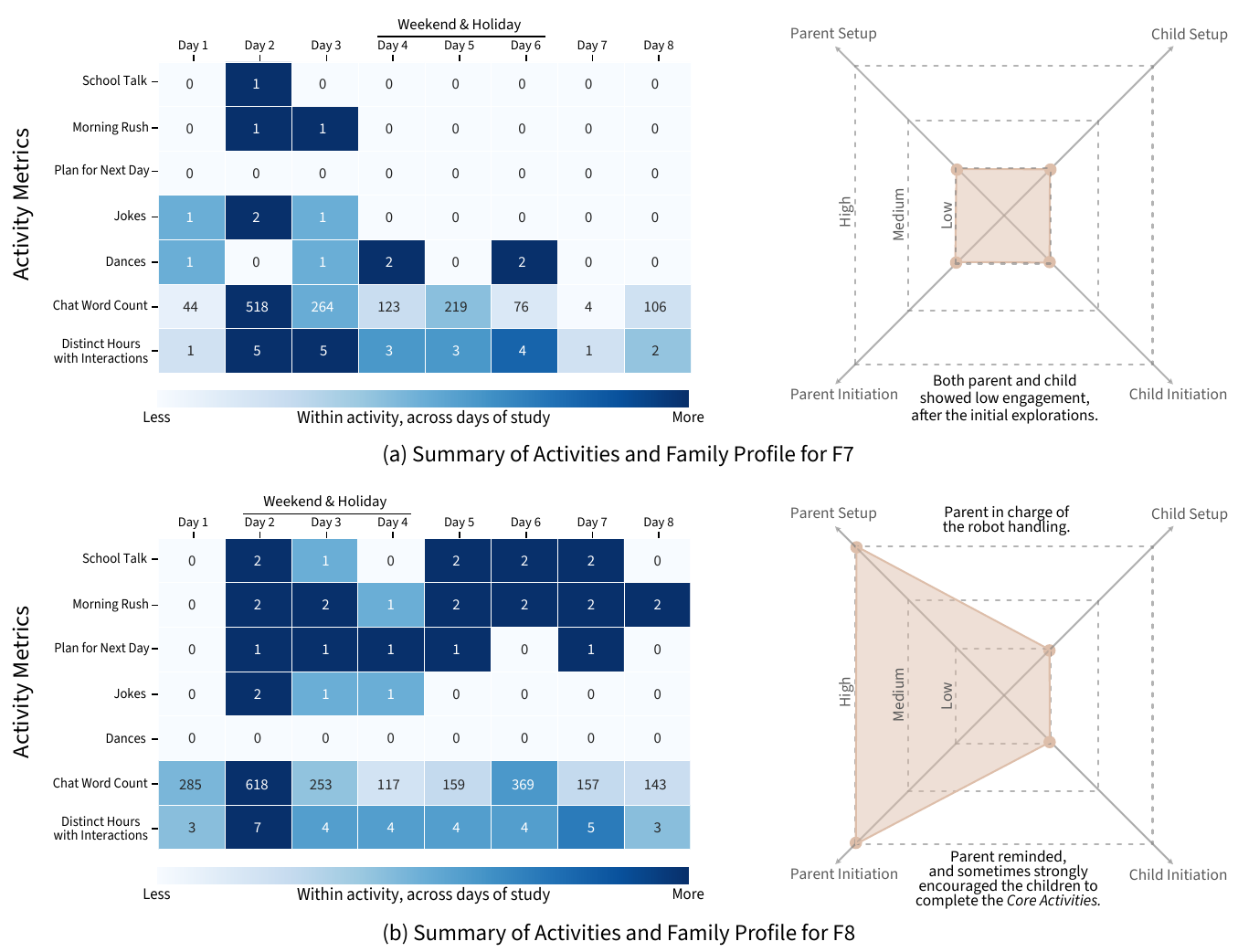}
  \caption{\textbf{Usage patterns from two other families.} (a) F7 (Low Engagement) and (b) F8 (Strongly Parent-led). For F7, there are only three instances of \textit{Core Activity} interactions, and sporadic interactions with the auxiliary features that also gradually decreased over time. For F8, we notice very consistent \textit{Core Activity} utilization, even on weekends. This can in part be explained by the parent's strong position on having the children complete the activities.}
  \Description[Activity summaries and family profiles for Families 7 and 8]{Two-panel figure comparing robot use for Families 7 and 8 across eight study days. In each row, the left side is a heatmap-style summary with columns for Day 1 through Day 8 and rows for School Talk, Morning Rush, Plan for Next Day, Jokes, Dances, Chat Word Count, and Distinct Hours with Interactions. Darker shading indicates greater activity within each metric across days, and weekend or holiday days are marked above the timeline. The right side of each row is a family profile diagram showing relative levels of parent setup, child setup, parent initiation, and child initiation. For Family 7, activity is concentrated in the early days of the deployment, with little use later on. The profile indicates low engagement from both parent and child after the initial exploration period. For Family 8, activity is more sustained across the study, especially for School Talk and Morning Rush, with no dance activity and moderate use of jokes and planning. The profile indicates that the parent was primarily responsible for handling the robot and often reminded or strongly encouraged the children to complete the core activities. Together, the figure contrasts an early drop-off pattern with a more parent-driven pattern of continued engagement.}
  \label{fig:overtime_heatmap_others}
\end{figure*}

In Section~\ref{sec:study3_results} we examined the range of parental roles, and presented in Figure~\ref{fig:overtime_heatmap_typical} the usage patterns of a representative family. To capture the broader view of family dynamics, Figure~\ref{fig:overtime_heatmap_others} presents two additional patterns observed. In one case --- the \textit{strongly parent-led} family (F8) --- the children very consistently completed the core activities, but didn’t try much of the auxiliary features. Although the kids compared completing the activities to performing chores (``\textit{like cleaning your room,}'' as one of the children from F8 put it), the father did still find the experience valuable, as it surfaced aspects of their school life he would not have otherwise learned. While the children expressed a desire to engage with the robot on their own terms, this structured use did appear to support the overarching goal of school-related communication. At the other end of the spectrum, the parent in family F7 showed little interest in the robot’s core features. As a result, usage centered more on auxiliary and casual interactions, with minimal engagement in school-related activities. This contrast highlights the range of parental involvement styles that shaped how the robot was used and experienced.

So what \textit{is} the ideal role of a parent in the context of a social robot that serves purposes beyond companionship and entertainment? While the father in F8 was able to glean valuable insights, a longer-term perspective suggests that the more sustainable approach may lie somewhere between parenting styles with high parental control and high child autonomy. Most families seemed to arrive organically at a balanced approach -- one in which \textit{some} encouragement from parents helped initiate engagement, while still allowing children the autonomy to guide and sustain their own interaction with the robot. The dosage of the encouragements varies across the families, though, with roughly half the families reporting an overall pattern of child-led usage and the other half showing more parent-led patterns. This observation aligns with prior work on parental involvement in learning contexts, which has shown that parents' time, energy, and facilitation styles significantly shape how children engage with interactive learning systems~\cite{ho2025set}. These patterns also resonate with literature on parental scaffolding, which describes how parents structure children's participation, provide guidance, and calibrate support during learning-related activities~\cite{mermelshtine2017parent, moe2018scaffolding}. Seen in this light, parents in our study were not merely facilitating access to the robot, but actively shaping how children engaged with it and what kinds of support were considered appropriate.
Together, these findings highlight the need for designs that balance parent facilitation with opportunities for child-led interaction -- flexible enough to adapt to diverse family dynamics and the various developmental stages of the children, each calling for a potentially different level of scaffolding.

\subsubsection{Children’s Receptiveness to the Robot}
Some families drew comparisons between the robot’s interactions with the children and their own parent–child exchanges, noting that children often appeared more receptive to the robot. As the mother from F9 explained, ``\textit{I just felt like they were more receptive to her (Misty) speaking than when I speak. Whenever I say [something], they just sort of [go] in one ear and out the other.}'' Other parents also echoed the utility of the robot as a neutral communicator that the children are more responsive to: ``\textit{Misty is almost sort of a neutral third party telling the kids `hey, it's time to do X,' whereas if it's one of us doing it, like in any family, they're like, `I don't want to do that,' or `Can I have five minutes?' So if it's not one of the parents telling them it's time to do it, they're more likely to do it}'' (P11). This perceived advantage extended beyond instructions to conversational exchanges. Even when the robot’s \textit{School Talk} prompts were heavily informed by parental input, some parents felt the robot elicited richer responses than if they had asked the same questions themselves. For example, as the father of F14 reflected, ``\textit{I just don't feel like it would be the same if I asked those [questions]. Well, I know it's not the same if I asked those same questions about, `what do you eat?' <The child makes silly sound in response to that question> Yeah, you know what I'm saying?}'' Although we did not specifically set out to investigate this aspect of the child-robot dynamic, these accounts seem to indicate a pattern that may warrant more systematic examination in future studies. Does the pattern persist in a larger-scale study? What characteristics of the children and the robot mediate the receptiveness? And more importantly, what are the implications for the ethical and responsible application of this dynamic in designing future child-robot interactions?

\subsection{Ethical and Privacy Considerations}
\label{sec:ethic-privacy}
The importance of ethics and privacy considerations has been consistently highlighted in prior works on robots in family contexts~\cite{levinson2024our, pearson2020child, cagiltay2020investigating}. To address these risks, we integrated procedural safeguards with specific technical design decisions. Our procedural approach involved informing families on data security and clarifying system behaviors (\textit{e.g.}, what data is captured and when). Complementing these, our technical design choices prioritized information security, including the provision of a secure local network separate from the family's own network, and we also provided a physical cover (\textit{i.e.}, a non-transparent tape) for the families to block the robot's camera whenever they desire. These measures ensured that families had substantive understanding and control over the robot's access to their data.

\subsubsection{The Use of LLM}
While our measures provided a baseline of control, the use of LLM, especially with young children, always warrants a careful consideration of the trade-offs. The first such trade-off involves commercial \textit{vs.} self-hosted LLMs. Commercial services provide researchers access to more sophisticated models that enable higher quality child-robot interactions, but self-hosted options provide more data agency (despite the security guarantees from the commercial options). In our study, we opted for the commercial option for three main reasons: (1) We did not expect the interactions to involve overly sensitive conversations; (2) A higher quality model can afford more natural interactions and better user experiences, lending more validity to our findings in a future-oriented perspective, and; (3) Errors and hallucinations from lower quality models themselves are harmful to the children and families as well. Given the rapid evolution of open-source LLMs capable of running on less demanding hardware, transitioning to self-hosted models represent an increasingly practical strategy that we are actively monitoring and evaluating. However, even with self-hosted models, the use of LLMs with children still requires careful evaluation of the risk and benefits~\cite{sun2026ai}. In our study, during the nightly manual check, the research team also examined the conversations between the child and the robot to ensure the safety of the children involved.

\subsubsection{Family Relational and Children Developmental Impact}
In addition to the immediate emotional bonds and heightened receptiveness discussed in Section~\ref{sec:parent-child-robot}, the integration of robots into the domestic sphere raises long-term concerns regarding psychological development and the evolution of caregiver-child interactions. While there is currently a lack of understanding in the field on the precise long-term impact these conversational robots may have on families, findings from other in-home technologies indicate that they could both foster and diminish relationships between parents and children~\cite{shin2021designing, padilla2012getting}. Our findings indicate that children often provide richer, more elaborate responses to the robot than to their parents. While this can serve as a functional tool for eliciting information or encouraging reflection, it does introduce a potential risk of displacing essential caregiver-child interactions. From a developmental perspective, the primary concern is whether a robot might inadvertently replace the communicative labor traditionally performed by parents. For example, if children increasingly turn to the robot for conversational validation or as a primary outlet for sharing their daily experiences, the frequency and depth of spontaneous parent-child exchanges may diminish. While activities in our study were designed to facilitate parent-child communications, and generally expect the parents to be around during the child-robot interactions, future work should examine how these dynamics shift, and more closely assess whether these systems act as a supplement that scaffolds family communication or a substitute that fundamentally alters the nature of the caregiver-child bond.

\subsection{Design Implications}
Building on our findings, we identify five main implications for designing social robots for family use. While grounded in the context of this study, these implications point to broader considerations for designing social robots that can integrate effectively into the complex social and physical environments of family life.

\textbf{1. Leverage the Robot as a ``Neutral Communicator.''} 
The findings suggest that the robot's value was not just in organizing data, but also in changing the social dynamic of communication and compliance. Children were more receptive to instructions from the robot than from their parents. This observation aligns with prior work on robot-supported routines~\cite{chan2017wakey, xu2026designing}, while extending it by highlighting the broader potential for robots to facilitate in high-friction tasks and activities. By offloading such interactions to the robot, these designs may reduce parent-child tension and allow parents to focus their efforts on relational and emotionally supportive aspects of family interaction.

Furthermore, this neutral stance facilitated a form of ``osmotic learning'' for parents, allowing them to gain insights into their children's school lives. Interaction designs should therefore explicitly \textbf{support passive parental awareness}, for instance by situating interactions in shared family spaces. In doing so, the robot functions as a ``window'' into the child's life, serving as a conversational catalyst that elicits richer responses than parents feel they can achieve on their own, while keeping the parent informed without requiring them to be excessively probing.

\textbf{2. Improve Context and State Awareness.} Families noticed when reminders or prompts were poorly timed (\textit{e.g.}, telling a child to pack lunch while still eating breakfast, or providing reminders to children who are not yet awake). These moments reduced the perceived usefulness of the robot and occasionally disrupted family routines. Prior work on reminders have also argued for the the utilization of more contextual information~\cite{dey2000cybreminder}. Recent development in and applications of Vision Language Models has shown potentials in allowing systems to better understand the context within which they are embedded~\cite{lim2024exploring,fang2025mirai}, including robotic systems used for contextual reminders~\cite{xu2026designing}, and may be a promising route to explore. For example, the robot may detect whether a child is awake, present, or already engaged in a relevant activity, which would make robot interventions more timely, relevant, and less intrusive. Such context-sensitivity specifically addresses parental feedback calling for interactions to emerge more organically---for example, by proactively asking about school-related topics in the afternoon, rather than relying on explicit activity switches and activations.

\textbf{3. Design for Multi-User Conflict and Negotiation.} As highlighted by \citet{cagiltay2024toward}, family-robot interaction is more than the sum of individual interactions with the robot. In our study, shared household contexts created both opportunities and challenges. Siblings sometimes had to take turns and occasionally fought over the robot (F6, F9, F10, F12). Family members not directly involved in the interaction could also influence use -- for instance, the mother from F8 occasionally turned off the robot because its reminders, while relevant to others, became distracting for her due to its placement nearby. These findings underscore the need for designers to consider not only the target user, but also how other household members perceive and interact with the robot.

\textbf{4. Reduce Friction via Ecosystem Integration.} While families perceived high value in the robot's organization capabilities, friction in setup and maintenance threatened long-term adoption. Families noted that small inconveniences, such as manually entering and updating schedule data, became barriers to sustained use. To support the busy and dynamic nature of family life, social robots cannot function as isolated silos. Instead, designs should aim for ``deep integration'' with the family's existing digital ecosystems, such as automatically syncing with family member's existing digital calendars and school's Learning Management Systems (LMS). By automating the flow of information, the robot can \textbf{shift from being another device to manage into a seamless extension of the family's existing coordination infrastructure}. However, as discussed earlier, this connectivity must be balanced with user control. As \citet{levinson2024our} highlight, families actively negotiate what information is appropriate for robots to access, treating privacy as a collective ``business'' rather than an individual setting. Deep integration must therefore be paired with granular privacy controls and transparent data policies, so the robot respects family boundaries while streamlining daily logistics.

Beyond software integration alone, prior work on technologies in homes also points to the \textbf{social dimensions of domestication}~\cite{soraa2021social}. Although prior work on integrating domestic robots into family routines is still limited, speculative and preliminary studies hint that aligning interactions with existing household routines may reduce friction and support sustained engagement~\cite{xu2024robots, cagiltay2024supporting}. While other mechanisms may also be at play, our findings echo this pattern: \textbf{routine-anchored activities} such as \textit{Morning Rush} showed more stable usage over time than auxiliary features, suggesting that ecosystem integration should encompass not only data sources but also temporal and routine alignment.

\textbf{5. Design Relational Infrastructure to Support Utility.} Our findings suggest that families experienced a symbiotic relationship between the \textit{auxiliary activities} and the \textit{core activities}, but also point to a broader design lesson: in family settings, relational engagement may function as a precursor to utility rather than as a separate or secondary concern. Prior work on long-term robot adoption in domestic settings has identified both utilitarian value and hedonic appeal as essential ingredients for sustained use~\cite{de2016long}. Consistent with this, although usage of auxiliary features in our study decreased as the initial novelty wore off, families described these interactions as serving a critical role in helping children build an initial connection with the robot and in supporting the formation of a habit of engaging with it. In this sense, playful interactions should not be treated merely as add-ons, but as \textbf{relational infrastructure that helps create the social conditions under which more functional activities can be taken up and sustained}. By fostering rapport, familiarity, and a sense that the robot is approachable and worth attending to, such interactions may make children more receptive to the robot's subsequent educational or organizational interventions. More broadly, these findings suggest that the usefulness of a family robot cannot be understood only in terms of what functions it provides, but also in terms of how it becomes socially meaningful enough for family members to listen to, respond to, and incorporate into everyday routines.

\subsection{Limitations}

While our findings offer rich and ecologically grounded preliminary insights into the design and the potential of a social robot to support home-based family-school communications, several limitations should be noted in terms of the extent of their generalizability. \textit{First}, our study involved a relatively small and demographically limited sample of ten to fourteen families. \textit{Second}, the families in the Evaluation Study also participated in the earlier design phases. Although this iterative process strengthened the family-centered fit of the prototype, it may also have shaped participants' perceptions of its usefulness. Future work with a new cohort of families is needed to more independently evaluate the system.
\textit{Third}, the one-week deployment period may not reflect how the robot would fare in sustained, longer-term use. \textit{Fourth}, limitations in the current prototype's capabilities -- such as reliance on parents' pre-deployment input and general latency issues, or errors in speech recognition -- may have influenced usage patterns and user perceptions. \textit{Fifth}, we did not explicitly examine how caregiving and school-related coordination were distributed within each household. Future work could more directly investigate how unequal care work shapes who engages with the robot, for what purposes, and with what expectations. Another limitation is that, although children were active participants in the study, many of the findings presented here still draw primarily from parent accounts, likely in part because parents tended to be more verbally expressive during the interviews. Future work should more deliberately foreground children's own perspectives on the robot's interaction style, usefulness, and role in family life. \textit{Finally}, the absence of direct teacher participation means the system was evaluated only on the family side of the partnership, leaving open questions about how it might integrate into the full family-school communication loop.

\subsection{Toward Direct School Integration}
While our framing is motivated by the broader concept of family-school partnerships, the prototype in this paper primarily supports the family side of that partnership rather than direct communication with schools or teachers. We view this as an important foundational layer of it. Family-school partnerships are sustained not only through formal school-facing channels, but also through the everyday work families do at home to interpret school information, coordinate schedules, remind children of responsibilities, and create opportunities for conversation about school experiences. Our findings suggest that a social robot may help scaffold these home-based practices, especially around organization, planning, and parent-child communication. In this sense, the present work lays groundwork for future systems that more directly connect with teachers, school platforms, or classroom activities, while emphasizing that successful school integration may first require supporting the family routines through which school-related information is actually taken up in everyday life.

A natural extension of the current project is to build on the home-based approach by incorporating teacher input into the robot’s activities and adding a feedback channel from families to teachers. Families felt that even without modifying the existing core activities, teacher contributions could significantly enrich the robot’s prompts. For example, almost all families expressed that academic check-ins from the teacher -- such as what the class is currently learning -- could feed into the \textit{School Talk} activity. Instead of the generic, ``What did you learn in math class today?'', the robot could ask more targeted questions, such as, ``Do you remember that cool thing we learned about multiplication?''

To close the loop of the family–school partnership, several families also discussed a variation of \textit{School Talk} that would serve as a backchannel to the school. In the current design, parents can receive a summary of their child’s \textit{School Talk} conversation; they suggested extending this so that a version could also be shared with teachers -- for example, to flag academic challenges or other concerns that might not surface in class.

Parents and children, however, raised concerns about the content of such summaries. Some preferred an approval process before sending information to the school, while others worried this would become tedious. Potential solutions included aggregating and anonymizing summaries across students, or creating a dedicated core activity that makes clear, ``some elements of this conversation may be shared with your teacher,'' with strict boundaries on what would be shared. These mechanisms could help address privacy concerns while maintaining transparency and trust.

\section{Conclusion}

Through a family-centered design process and a week-long in-home evaluation, this work demonstrates the potential of social robots to meaningfully support home-based activities that facilitate school-related communication. Drawing on formative design work and in-home use, we identified a core set of robot-facilitated activities, including evening planning, morning reminders, and school-related conversations. Our findings highlight that there is no ``one-size-fits-all'' approach: successful adoption depends on accommodating diverse family preferences. We further observed that integration into daily life is strongly shaped by existing family dynamics, particularly parental facilitation styles and sibling interactions. Beyond these patterns, we provide preliminary evidence that the robot can facilitate timely information exchange and reduce parental burden by offloading specific logistical tasks. Synthesizing these insights, we outline actionable design considerations that emphasize playful engagement as a scaffold for utility, deep integration with family ecosystems, and sensitivity to contextual and social dynamics. Looking forward, incorporating teacher input and feedback channels could extend these benefits and support more complete family-school partnerships.

\section*{Selection and Participation of Children}
The study protocol was reviewed and approved by the authors’ Institutional Review Board. Families were recruited through the University of Wisconsin--Madison research recruitment mailing list, specifically the segment consisting of faculty and staff members. The inclusion criterion was having at least one child between the ages of 9 and 12 who lives with them, fluent in English, and lives within the Madison metropolitan area (WI, USA). Siblings outside this target age range were also encouraged to participate. A total of 14 families initially enrolled in the research, with 10 families completing the full sequence of three studies. Participating children predominantly attended local public schools. The demographic profile of the households was primarily middle- to upper-income, with the majority of families self-reporting annual household incomes of \$100,000 USD or higher.

Parents provided written informed consent for themselves and their children prior to participation, and children provided age-appropriate verbal assent. Families were informed of study procedures, including the data collection involved, and were reminded of their right to withdraw at any time. Families consented to and were compensated for each study phase independently. For the needs-finding interviews and co-design sessions, families received \$20 USD per hour. For the final week-long in-home evaluation, families were provided a flat payment of \$50 USD upon completion. 

As detailed in Section~\ref{sec:ethic-privacy}, we adopted both procedural safeguards and technical design decisions to uphold data security and privacy. In addition, the three-phase study design allowed children to observe how their input contributed to shaping the developed technology, while giving families time to become familiar with the system, particularly with respect to what data were collected in different situations. This also provided ample opportunities for families to raise questions or concerns prior to the in-home evaluation.

\section*{Supplementary Materials}
Supplementary Materials are accessible \href{https://osf.io/fcqr4}{here}\footnote{\url{https://osf.io/fcqr4/}}. Included within are: (1) Anonymized interview transcripts for all three studies; (2) Visualized usage patterns (similar to Figure~\ref{fig:timeline1}) for all ten families; (3) More background information on the participating families; (4) The instruction booklets, one intended for parents and one for children; (5) Example code snippets illustrating the LLM prompts and the main logic driving the interactions, and; (6) More technical details regarding the system.

\begin{acks}
This work was supported by the National Science Foundation award \#2312354. Figures~\ref{fig:teaser} and~\ref{fig:example_main_activities} used vector art assets from Freepik~\cite{freepik}. 
\end{acks}

\bibliographystyle{ACM-Reference-Format}
\balance
\bibliography{references}


\end{document}